%% file: Main_MoTra21.tex
\documentclass[ctexart, utf8, 11pt]{article}
\usepackage{nodalida2021}
\usepackage{times}
\usepackage{soul}
\usepackage{epstopdf}
\usepackage{booktabs}

\usepackage{breqn}
\usepackage{amsmath}
\usepackage{comment}
\usepackage{graphicx}
\usepackage{CJKutf8}  % added by Aaron
\usepackage[utf8]{inputenc}
\usepackage[T1]{fontenc}
\usepackage{graphicx,amsmath,amsfonts,amssymb,amscd,amsthm}
\usepackage{url}
\usepackage{hyperref} %added by Aaron
\usepackage{amsmath}
\usepackage{latexsym}
\usepackage{pgfplots}
\pgfplotsset{compat=1.14}
\usepackage{pgfplotstable}

\definecolor{aqua}{rgb}{0.0, 1.0, 1.0}

\newlength\graphheight
\newlength\graphwidth
\pgfplotsset {
graph/.style={
height=\graphheight,
width=\graphwidth,
legend style={
    draw=none,
    fill=none,
    font=\footnotesize,
},
ticklabel style={font=\scriptsize, /pgf/number format/fixed},
scaled y ticks = false,
scaled x ticks = false,
xlabel near ticks,
ylabel near ticks,
},
graphright/.style={
graph,
yticklabel pos=right,
},
}

\aclfinalcopy % Uncomment this line for the final submission
\title{Translation Quality Assessment: A Brief Survey on Manual and Automatic Methods}

\author{
           Lifeng Han$^1$, Gareth J. F. Jones$^1$, \and Alan F. Smeaton$^2$ \\

         $^1$ ADAPT Research Centre  \\ 
         $^2$ Insight Centre for Data Analytics 
         \\ School of Computing, Dublin City University, Dublin, Ireland \\
         {\tt lifeng.han@adaptcentre.ie} \\}
         
\date{}

\begin{document}

\maketitle

\begin{abstract}
To facilitate effective translation modeling and translation studies, one of the crucial questions to address is how to assess  translation quality. From the perspectives of accuracy, reliability, repeatability and cost,  translation quality assessment (TQA) itself is a rich and challenging task. In this work, we present a high-level and concise survey of TQA methods, including both manual judgement criteria and automated evaluation metrics, which we classify into further detailed sub-categories. We hope that this work will be an asset for both translation model researchers and quality assessment researchers. In addition, we hope that it will enable practitioners to quickly develop a better understanding of the conventional TQA field, and to find corresponding closely relevant evaluation solutions for their own needs. This work may also serve inspire further development of quality assessment and evaluation methodologies for other natural language processing (NLP) tasks  in addition to machine translation (MT), such as automatic text summarization (ATS), natural language understanding (NLU) and natural language generation (NLG). \footnote{authors GJ and AS in alphabetic order}
\end{abstract}

% \begin{CJK*}{UTF8}{gbsn} 中学 \end{CJK*}
%\begin{CJK} 中学\end{CJK}  

\input{sec-intro}

\input{sec-Human-TQA}

\input{sec-Auto-TQA}

%to-add-back \input{sec-Evaluate-TQA}

\input{sec-perspective}

\section{Conclusions and Future Work}

In this paper we have presented a survey of the state-of-the-art in translation quality assessment methodologies from the viewpoints of both manual judgements and automated methods. This work differs from conventional MT evaluation review work by its concise structure and inclusion of some recently published work and references.
Due to space limitations, in the main content, we focused on conventional human assessment methods and automated evaluation metrics with reliance on reference translations. However, we also list some interesting and related work in the appendices, such as the quality estimation in MT when the reference translation is not presented during the estimation, and the evaluating methodology for TQA methods themselves.
However, this arrangement does not affect the overall understanding of this paper as a self contained overview. 
We believe this work can help both MT and NLP researchers and practitioners in identifying appropriate quality assessment methods for their work. We also expect this work might shed some light on  evaluation methodologies in other NLP tasks, due to the similarities they share, such as text summarization \cite{Mani2001SummarizationEA,bhandari-etal-2020-evaluating_summarization}, natural language understanding \cite{2021NLU_ruder_multilingual_evaluation}, 
natural language generation \cite{arxiv2021NLG}, as well as programming language (code) generation \cite{code_generation2021arXiv}.

%language generation, language understanding and automatic summarization \cite{arxiv2021NLG,2021NLU_ruder_multilingual_evaluation,Mani2001SummarizationEA,bhandari-etal-2020-evaluating_summarization}

\section*{Acknowledgments}

We appreciate the comments from Derek F. Wong, editing help from Ying Shi (Angela), and the anonymous reviewers for their valuable reviews and feedback.
The ADAPT Centre for Digital Content Technology is funded under the SFI Research Centres Programme (Grant 13/RC/2106) and is co-funded under the European Regional Development Fund. The input of Alan Smeaton is part-funded by Science Foundation Ireland 
under grant number SFI/12/RC/2289 (Insight Centre). 
%Do not number the acknowledgment section. Do not include this section when submitting your paper for review.

\bibliographystyle{acl_natbib}
\bibliography{nodalida2021}

\input{Appendix}

\end{document}

%% file: sec-intro.tex
\section{Introduction}
%to choose: 
Machine translation (MT) research, starting from the 1950s  \cite{Weaver1955}, has been one of the main research topics in computational linguistics (CL) and natural language processing (NLP), and has influenced and been influenced by several other language processing tasks such as parsing and language modeling. Starting from rule-based methods to example-based, and then statistical methods \cite{brown-etal-1993-mathematics,Och2003CL,chiang-2005-hierarchical,Koehn2010}, to the current paradigm of neural network structures \cite{cho2014encoder-decoder,Google2016MultilingualNMT,google2017attention,bert4mt2019lample}, MT quality continue to improve. However, as MT and translation quality assessment (TQA) researchers report,  MT outputs are still far from reaching human parity \cite{laubli2018mt_human_parity,L_ubli_2020_human_parity,han-etal-2020-alphamwe}. MT quality assessment is thus still an important task to facilitate MT research itself, and also for downstream applications. TQA remains a challenging and difficult task because of the richness, variety, and ambiguity phenomena of natural language itself, e.g. the same concept can be expressed in different word structures and patterns in different languages, even inside one language \cite{Arnold2003}.

In this work, we introduce  human judgement and evaluation (HJE) criteria that have been used in standard international shared tasks and more broadly, such as NIST \cite{Li2005}, WMT \cite{koehnMonz2006,BurchFordyceKoehnMonz2007,BurchKoehnMonzSchroeder2008,BurchKoehnMonzSchroeder2009,BurchKoehnMonzPeterson2010,BurchKoehnMonzZaidan2011,BurchKoehnMonzPostSoricut2012,BojarBuckBurchFedermann2013,bojar-EtAl-2014-W14-33,bojar-EtAl-2015-WMT,bojar-etal-wmt_metrics2016-results,wmt2017findings,wmt2018findings,barrault-etal-wmt2019-findings,barrault-etal-wmt2020_findings}, and IWSLT \cite{EckHori2005,Paul2009,PaulFedericoStuker2010,FedericoBentivogliPaulStuker2011}. We then introduce  automated TQA methods, including the automatic evaluation metrics that were proposed inside these shared tasks and beyond.
Regarding Human Assessment (HA) methods, we categorise them into traditional and advanced sets, with the first set including intelligibility, fidelity, fluency, adequacy, and comprehension, and the second set including task-oriented, extended criteria, utilizing post-editing, segment ranking, crowd source intelligence (direct assessment), and revisiting traditional criteria. 

Regarding automated TQA methods, we classify these into three categories including simple n-gram based word surface matching, deeper linguistic feature integration such as syntax and semantics, and deep learning (DL) models, with the first two regarded as traditional and the last one regarded as advanced due to the recent appearance of DL models for NLP. We further divide each of these three categories into sub-branches, each with a different focus. Of course, this classification does not have clear boundaries. For instance some automated metrics are involved in both n-gram word surface similarity and linguistic features. This paper differs from the existing works \cite{GALEprogram2009,EuroMatrixProject2007} by introducing recent developments in MT evaluation measures, the different classifications from manual to automatic evaluation methodologies, the introduction of more recently developed quality estimation (QE) tasks, and its concise presentation of these concepts.

We hope that our work will shed light and offer a useful guide for both MT researchers and researchers in other relevant NLP disciplines, from the similarity and evaluation point of view, to find useful quality assessment methods, either from the manual or automated perspective, inspired from this work. This might include, for instance, natural language generation \cite{arxiv2021NLG}, natural language understanding \cite{2021NLU_ruder_multilingual_evaluation},  and automatic summarization \cite{Mani2001SummarizationEA,bhandari-etal-2020-evaluating_summarization}.%, and searching \cite{search_eval2021Liu}.

%NLG_evaluation2017emnlp_Novikova_etal

%language generation, language understanding and automatic summarization \cite{arxiv2021NLG,2021NLU_ruder_multilingual_evaluation,Mani2001SummarizationEA,bhandari-etal-2020-evaluating_summarization}.

%such as natural language generation \cite{arxiv2021NLG}, natural language understanding \cite{2021NLU_ruder_multilingual_evaluation} and automatic summarization \cite{Mani2001SummarizationEA,bhandari-etal-2020-evaluating_summarization}, due to the similarities they share.

%The rest of the paper is organized as follows: Section 2 and 3 present the human assessment and automated assessment methods respectively; section 4 comes the evaluating methods of the quality assessment models; Section 5 introduces advanced MT quality estimation (QE); Section 6 discusses and concludes the paper with perspectives.

%The rest of the paper is organized as follows: Section 2 and 3 present the human assessment and automated assessment methods respectively; section 4 comes the evaluating methods of the quality assessment models; Section 5 concludes the paper; and we list some further relevant reading into appendices, such as MT quality estimation (QE). %introduces advanced MT quality estimation (QE); Section 6 discusses and concludes the paper with perspectives.

The rest of the paper is organized as follows: Sections 2 and 3 present human assessment and automated assessment methods respectively;
Section 4 presents some discussions and perspectives;  Section 5 summarizes our conclusions and future work.  We also list some further relevant readings in the appendices, such as evaluating methods of TQA itself, MT QE, and mathematical formulas.\footnote{This work is based on an earlier preprint edition \cite{han2016MTE_survey}} 
%introduces advanced MT quality estimation (QE); Section 6 discusses and concludes the paper with perspectives.
%han2016MTE_survey

%% file: sec-Human-TQA.tex
\section{Human Assessment Methods}

\begin{figure}[!t]
\centering
\includegraphics*[height=2.1in,width=3in]{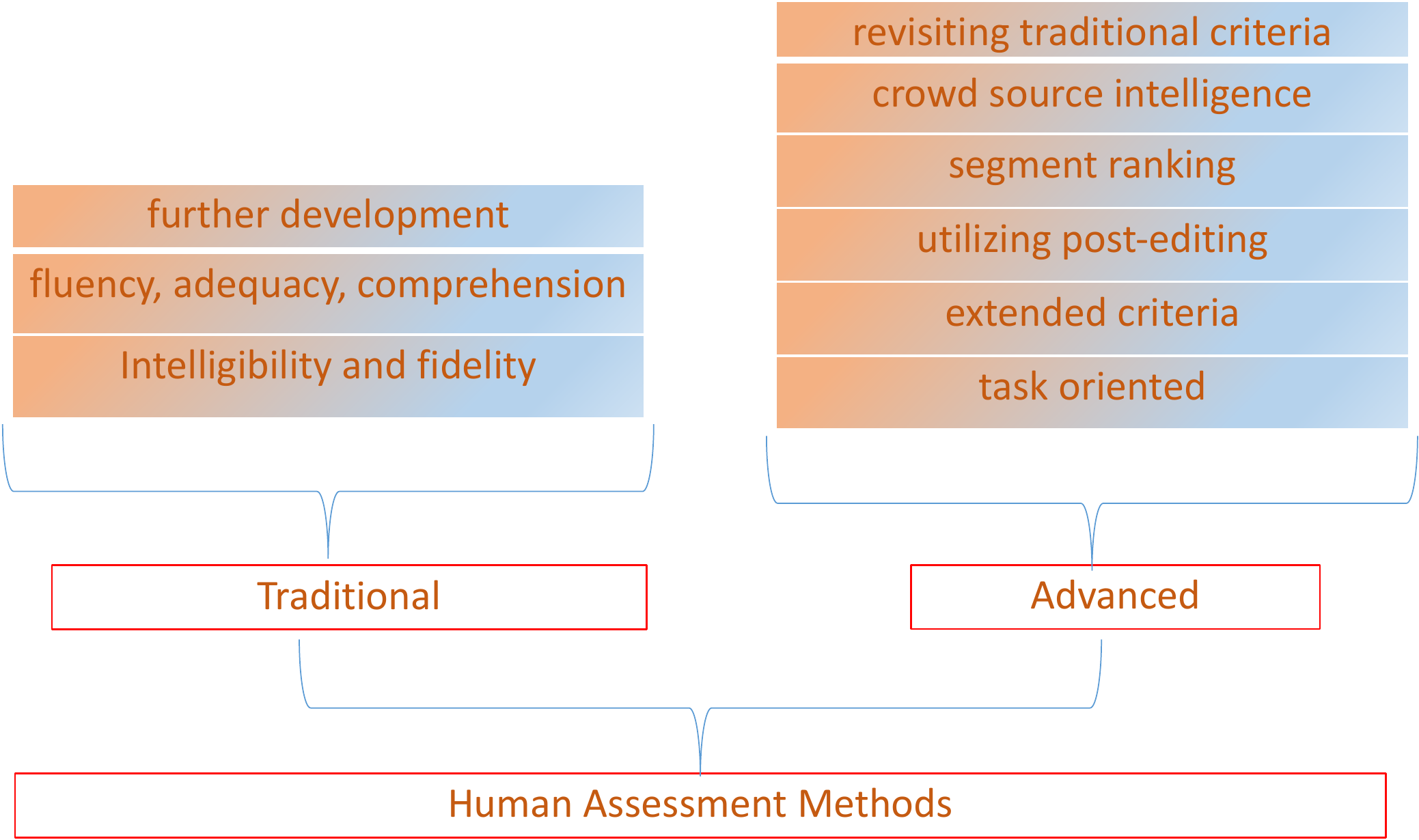}
%[height=3.1in,width=2.6in]
\caption{Human Assessment Methods}
\label{fig:humanMTEfig}
\end{figure}

In this section we introduce human judgement methods, as reflected in Fig.~\ref{fig:humanMTEfig}. This categorises these human methods as Traditional and Advanced.  

\subsection{Traditional Human Assessment}
\subsubsection{\emph{Intelligibility and Fidelity}}

% \textbf{}
The earliest human assessment methods for MT can be traced back to around 1966. They include the intelligibility and fidelity used by the automatic language processing advisory committee (ALPAC) \cite{Carroll1966}.
The requirement that a translation is intelligible means that, as far as possible, the translation should read like normal, well-edited prose and be readily understandable in the same way that such a sentence would be understandable if originally composed in the translation language. The requirement that a translation is of high fidelity or accuracy includes the requirement that the translation should, as little as possible, twist, distort, or controvert the meaning intended by the original.

\subsubsection{\emph{Fluency, Adequacy and Comprehension}}

In 1990s, the Advanced Research Projects Agency (ARPA) created a methodology to evaluate machine translation systems using the adequacy, fluency and comprehension of the MT output \cite{ChurchHovy1991} which adapted in MT evaluation campaigns including \cite{WhiteConnellMara1994}.

\begin{comment}
\begin{eqnarray}
\text{Comprehension}=\frac{\#\text{Cottect}}{6} \\
\text{Fluency}=\frac{\frac{\text{Judgment point}-1}{\text{S}-1}}{\#\text{Sentences in passage}} \\
\text{Adequacy}=\frac{\frac{\text{Judgment point}-1}{\text{S}-1}}{\#\text{Fragments in passage}} 
\end{eqnarray}
\end{comment}

To set upp this methodology, the human assessor is asked to look at each fragment, delimited by syntactic constituents and containing sufficient information, and judge its adequacy on a scale 1-to-5. Results are computed by averaging the judgments over all of the decisions in the translation set.

Fluency evaluation is compiled in the same manner as  for the adequacy except  that the assessor is to make intuitive judgments on a sentence-by-sentence basis for each translation. Human assessors are asked to determine whether the translation is good English without reference to the correct translation. Fluency evaluation determines whether a sentence is well-formed and fluent in context.

Comprehension relates to ``Informativeness'', whose objective is to measure a system's ability to produce a translation that conveys sufficient information, such that people can gain necessary information from it. The reference set of expert translations is used to create six questions with six possible answers respectively including, ``none of above'' and ``cannot be determined''.

\subsubsection{\emph{Further Development}}

\newcite{BangaloreRambowWhittaker2000} classified  accuracy into several categories including simple string accuracy, generation string accuracy, and two corresponding tree-based accuracy. \newcite{reeder2004amta} found the correlation between fluency and the number of words it takes to distinguish between human translation and MT output.

The ``Linguistics Data Consortium (LDC)'' \footnote{https://www.ldc.upenn.edu} designed two five-point scales representing fluency and adequacy for the annual NIST MT evaluation workshop. The developed scales became a widely used methodology when manually evaluating MT by assigning values. The five point scale for adequacy indicates how much of the meaning expressed in the reference translation is also expressed in a translation hypothesis; the second five point scale indicates how fluent the translation is, involving both grammatical correctness and idiomatic word choices. 

\newcite{SpeciaHajlaouiHallettAziz2011} conducted a study of  MT adequacy and broke it into four levels, from score 4 to 1: highly adequate, the translation faithfully conveys the content of the input sentence; fairly adequate, where the translation generally conveys the meaning of the input sentence, there are some problems with word order or tense/voice/number, or there are repeated, added or non-translated words; poorly adequate, the content of the input sentence is not adequately conveyed by the translation; and completely inadequate, the content of the input sentence is not conveyed at all by the translation.

\subsection{Advanced Human Assessment}

\subsubsection{\emph{Task-oriented}}

\newcite{WhiteTaylor1998} developed a task-oriented evaluation methodology for Japanese-to-English translation to measure MT systems in light of the tasks for which their output might be used. They seek to associate the diagnostic scores assigned to the output used in the DARPA (Defense Advanced Research Projects Agency) \footnote{https://www.darpa.mil} evaluation with a scale of language-dependent tasks, such as scanning, sorting, and topic identification. They  develop an MT proficiency metric with a corpus of multiple variants which are usable as a set of controlled samples for user judgments. The principal steps include identifying the user-performed text-handling tasks, discovering the order of text-handling task tolerance, analyzing the linguistic and non-linguistic translation problems in the corpus used in determining task tolerance, and developing a set of source language patterns which correspond to diagnostic target phenomena. A brief introduction to task-based MT evaluation work was shown in their later work \cite{Doyon99task-basedevaluation}.

\newcite{Voss06task-basedevaluation} introduced tasked-based MT output evaluation by the extraction of \textit{who}, \textit{when}, and  \textit{where} three types of elements. They extended their work later into event understanding  \cite{Laoudi06task-basedmt}.

\subsubsection{\emph{Extended Criteria}}

\newcite{KingbelisHovy2003} extend a large range of manual evaluation methods for MT systems which, in addition to the earlir mentioned accuracy, include \textit{suitability}, whether even accurate results are suitable in the particular context in which the system is to be used; \textit{interoperability}, whether with other software or hardware platforms; \textit{reliability}, i.e., don't break down all the time or take a long time to get running again after breaking down; \textit{usability}, easy to get the interfaces, easy to learn and operate, and looks pretty; \textit{efficiency}, when needed, keep up with the flow of dealt documents; \textit{maintainability}, being able to modify the system in order to adapt it to particular users; and \textit{portability}, one version of a system can be replaced by a new version, because MT systems are rarely static and they tend to  improve over time as resources grow and bugs are fixed.

\subsubsection{\emph{Utilizing Post-editing}}

One alternative method to assess MT quality is to compare the post-edited correct translation to the original MT output. This type of evaluation is, however, time consuming and depends on the skills of the human assessor and post-editing performer. One example of a metric that is designed in such a manner is the human translation error rate (HTER) \cite{SnoverDorrSchwartzMicciulla2006}. This is based on the number of editing steps, computing the editing steps between an automatic translation and a reference translation. Here, a human assessor has to find the minimum number of insertions, deletions, substitutions, and shifts to convert the system output into an acceptable translation. HTER is defined as the sum of the  number of editing steps divided by the number of words in the acceptable translation.

\subsubsection{\emph{Segment Ranking}}

In the WMT metrics task,  human assessment based on segment ranking was often used. Human assessors were frequently asked to provide a complete ranking over all the candidate translations of the same source segment \cite{BurchKoehnMonzZaidan2011,BurchKoehnMonzPostSoricut2012}. In the WMT13 shared-tasks \cite{BojarBuckBurchFedermann2013}, five systems were randomised for the assessor to give a rank. Each time, the source segment and the reference translation were presented together with the candidate translations from the five systems. The assessors ranked the systems from 1 to 5, allowing tied scores. For each ranking, there was the potential to provide as many as 10 pairwise results if there were no ties. The collected pairwise rankings were then used to assign a corresponding score to each participating system to reflect the quality of the automatic translations. The assigned scores could also be used to reflect how frequently a system was judged to be better or worse than other systems when they were compared on the same source segment, according to the following formula:

\begin{small}
\begin{eqnarray}
\frac{\#\text{better pairwise ranking}}{\#\text{total pairwise comparison}-\#\text{ties comparisons}}
\end{eqnarray}
\end{small}

%\begin{equation} \frac{4}{5}\end{equation}
% \begin{equation} \frac{better_pairwise_rank}{total_pairwise_comparison-_ties_comparison}\end{equation}

\subsubsection{\emph{Crowd Source Intelligence}}

With the reported very low human inter-agreement scores from the WMT segment ranking task, researchers started to address this issue by exploring new human assessment methods, as well as seeking reliable automatic metrics for segment level ranking \cite{DBLP:conf/naacl/GrahamBM15}. 

\newcite{graham-etal-2013-continuous} noted that the lower agreements from WMT human assessment might be caused partially by the 
interval-level scales set up for the human assessor to choose regarding quality judgement of each segment. For instance, the human assessor possibly corresponds to the situation where neither of the two categories they were forced to choose is preferred. In light of this rationale, they proposed continuous measurement scales (CMS) for human TQA using fluency criteria. This was implemented by introducing the crowdsource platform Amazon MTurk, with some quality control methods such as the insertion of \textit{bad-reference} and \textit{ask-again}, and statistical significance testing. This methodology reported improved both intra-annotator and inter-annotator consistency. Detailed quality control methodologies, including statistical significance testing were documented in direct assessment (DA) \cite{NLEDA2016,graham-etal-2020-statistical}.%to-add-back: which has been adopted by recent WMT shared tasks.

% DA 2014/2015 on?

\subsubsection{\emph{Revisiting Traditional Criteria}}
%Similarly, not only to avoid the low inter-agreement scores caused by choosing some pre-defined error types or assigning scores, but also to give the human assessors more flexible choices to reflect the real problem from candidate translation they are looking at, xxx desgined new TQA methodology by asking assessors to mark all problematic parts of the translation, including words, phrases and sentences. 

\newcite{popovic-2020coling-informative} criticized the traditional human TQA methods because they fail to reflect real problems in translation by assigning scores and ranking several candidates from the same source. Instead, \newcite{popovic-2020coling-informative} designed a new methodology by asking human assessors to mark all problematic parts of candidate translations, either words, phrases, or sentences. 
Two questions that were typically asked of the assessors related to \textit{comprehensibility} and \textit{adequacy}. The first criteria considered whether the translation is understandable, or understandable but with errors; the second criteria measures if the candidate translation has different meaning to the original text, or maintains the meaning but with errors. Both criteria take into account whether parts of the original text are missing in translation.
Under a similar experimental setup, \newcite{popovic-2020conll-relations} also summarized the most frequent error types that the annotators recognized as misleading translations.

%% file: sec-Auto-TQA.tex
\section{Automated Assessment Methods}

Manual evaluation suffers some disadvantages such as that it is time-consuming, expensive, not tune-able, and not reproducible. Due to these aspects, automatic evaluation metrics have been widely used for MT. Typically, these compare the output of MT systems against human reference translations, but there are also some metrics that do not use reference translations. There are usually two ways to offer the human reference translation, either offering one single reference or offering multiple references for a single source sentence  \cite{LinOch2004,HanWongChao2012}. 

\begin{figure}[!t]
\centering
\includegraphics*[height=2.5in,width=3in]{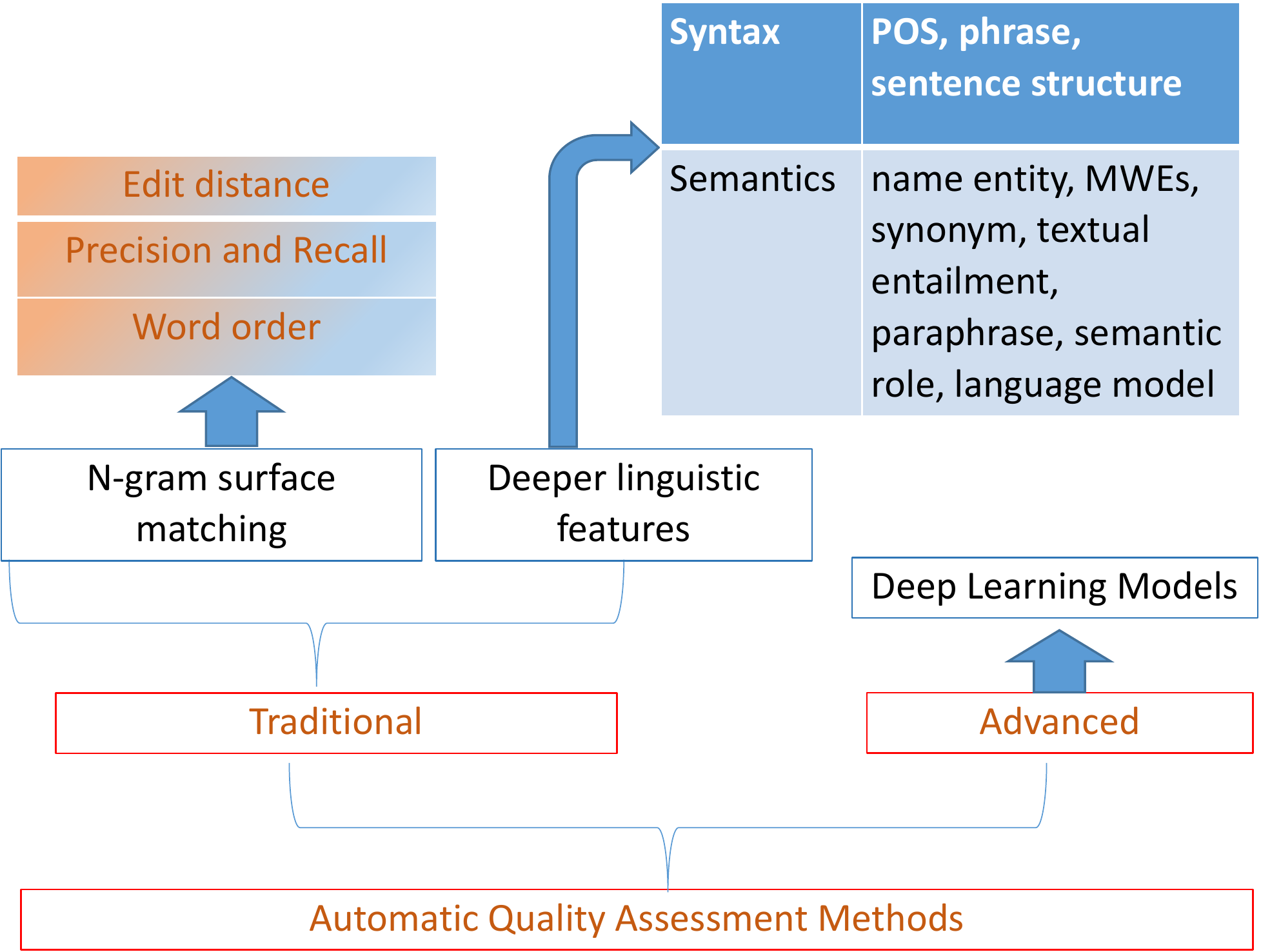}
%[height=3.1in,width=2.6in]
\caption{Automatic Quality Assessment Methods}
\label{fig:automaticMTEfig}
\end{figure}

Automated metrics often measure the overlap in words and word sequences, as well as word order and edit distance. We classify these kinds of metrics as ``simple n-gram word surface matching''. Further developed metrics also take linguistic features into account such as syntax and semantics, including POS, sentence structure, textual entailment, paraphrase, synonyms, named entities, multi-word expressions (MWEs), semantic roles and language models. We classify these metrics that utilize the linguistic features as ``Deeper Linguistic Features (aware)''. This classification is only for easier understanding and better organization of the content. It is not easy to separate these two categories clearly since sometimes they merge with each other. For instance, some metrics from the first category might also use certain linguistic features. Furthermore, we will introduce some recent models that apply deep learning into the TQA framework, as in Fig.~\ref{fig:automaticMTEfig}. Due to space limitations, we present MT quality estimation (QE) task which does not rely on reference translations during the automated computing procedure in the appendices.

\subsection{N-gram Word Surface Matching}
\subsubsection{\em{Levenshtein Distance}}

By calculating the minimum number of editing steps to transform MT output to reference, \newcite{SuWenShin1992} introduced the word error rate (WER) metric into MT evaluation. This metric, inspired by Levenshtein Distance (or edit distance), takes word order into account, and the operations include insertion (adding word), deletion (dropping word) and replacement (or substitution, replace one word with another), the minimum number of editing steps needed to match two sequences.

\begin{comment}
\begin{eqnarray}
\text{WER}=\frac{\text{substitution+insertion+deletion}}{\text{reference}_{\text{length}}}.
\end{eqnarray}

\end{comment}

%GJ: make clear why you are referring to Levenshtein distance here
One of the weak points of the WER metric is the fact that word ordering is not taken into account appropriately. The WER scores are very low when the word order of system output translation is ``wrong" according to the reference. In the Levenshtein distance, the mismatches in word order require the deletion and re-insertion of the misplaced words. However, due to the diversity of language expressions, some so-called ``wrong" order sentences by WER also prove to be good translations. To address this problem, the position-independent word error rate (PER) introduced by \newcite{TillmannVogelNeyZubiagaSawaf1997} is designed to ignore word order when matching output and reference. Without taking into account of the word order, PER counts the number of times that identical words appear in both sentences. Depending on whether the translated sentence is longer or shorter than the reference translation, the rest of the words are either insertion or deletion ones.
\begin{comment}
\begin{small}
\begin{equation}
\text{PER}
=1-\frac{\text{correct}-\max(0,\text{output}_{\text{length}}-\text{reference}_{\text{length}})}{\text{reference}_{\text{length}}}.
\end{equation}
\end{small}
\end{comment}

Another way to overcome the unconscionable penalty on word order in the Levenshtein distance is adding a novel editing step that allows the movement of word sequences from one part of the output to another. This is something a human post-editor would do with the cut-and-paste function of a word processor. In this light, \newcite{SnoverDorrSchwartzMicciulla2006} designed the translation edit rate (TER) metric that adds block movement (jumping action) as an editing step. The shift option is performed on a contiguous sequence of words within the output sentence. 
\begin{comment}
The TER score is calculated as:

\begin{equation} 
\text{TER}=\frac{\#\text{of edit}}{\#\text{of average reference words}}
\end{equation}

\end{comment}
For the edits, the cost of the block movement, any number of continuous words and any distance, is equal to that of the single word operation, such as insertion, deletion and substitution.

\subsubsection{\emph{Precision and Recall}}

The widely used evaluation BLEU metric \cite{PapineniRoukosWardZhu2002} is based on the degree of n-gram overlap between the strings of words produced by the MT output and the human translation references at the corpus level. BLEU calculates precision scores with n-grams sized from 1-to-4, together multiplied by the coefficient of brevity penalty (BP). If there are multi-references for each candidate sentence, then the nearest length as compared to the candidate sentence is selected as the effective one.
In the BLEU metric, the n-gram precision weight $\lambda_n$  is usually selected as a uniform weight. However, the 4-gram precision value can be very low or even zero when the test corpus is small. To weight more heavily those n-grams that are more informative, \newcite{Doddington2002} proposes the NIST metric with the information weight added.
\begin{comment}
\begin{eqnarray} 
\text{Info}=\log_{2}\big(\frac{\#\text{occurrence of}~w_1,\cdots,w_{n-1}}{\#\text{occurrence of}~ w_1,\cdots,w_n}\big)
\end{eqnarray}
\end{comment}
Furthermore, \newcite{Doddington2002} replaces the geometric mean of co-occurrences with the arithmetic average of $n$-gram counts, extends the $n$-gram into 5-gram ($N=5$), and selects the average length of reference translations instead of the nearest length.

%\begin{align}
%\text{NIST}&=\sum_{n=1}^{N}\{\frac{\sum\limits_{\text{all}~w_{1}\cdots{w}_{n}~\text{that coocur}}Info(w_{1}\cdots{w}_{n})}{\sum\limits_{\text{all}~w_{1}\cdots{w}_{n}\text{in sysoutput}}(1)}\}\nonumber\\
%&\times\exp\{\beta\log^{2}[\min(\frac{L_{sys}}{\bar{L_{ref}}},1)]\}
%\end{align}
%
%where $w_{1},\cdots,w_{n}$ means the words sequence, $\bar{L_{ref}}$ is the average number of words in a reference translation, averaged over all reference translations.

ROUGE \cite{LinHovy2003} is a recall-oriented evaluation metric, which was initially developed for summaries, and inspired by BLEU and NIST. %Following the adoption by the MT community of automatic evaluation using the BLEU/NIST scoring process, Lin conducts a study of a similar idea for evaluating summaries. 
ROUGE has also been applied in automated TQA in later work \cite{LinOch2004}. 

The F-measure is the combination of precision ($P$) and recall ($R$), which was firstly employed in  information retrieval (IR) and latterly adopted by the information extraction (IE) community, MT evaluations, and others.
\newcite{turian2006evaluation} carried out experiments to examine how standard measures such as precision, recall and F-measure can be applied to TQA and showed the comparisons of these standard measures with some alternative evaluation methodologies. 

\begin{comment}
\begin{eqnarray}
F_{\beta}=(1+\beta^{2})\frac{PR}{R+\beta^{2}P}
\end{eqnarray}

\end{comment}

%The variable $F_{\beta}$ measures the effectiveness of retrieval with respect to a user who attaches $\beta$ times as much importance to recall as precision.Traditional F-measure or balanced F-score ($F_{1}$ score) is exactly the harmonic mean of precision and recall, putting the same trade-off on precision and recall, $\beta=1$ , (Sasaki and Fellow, 2007).

\newcite{BanerjeeLavie2005} designed METEOR as a novel evaluation metric. METEOR is based on the general concept of flexible unigram matching,  precision and  recall, including the match between words that are simple morphological variants of each other with identical word stems and words that are synonyms of each other. To measure how well-ordered the matched words in the candidate translation are in relation to the human reference, METEOR introduces a penalty coefficient, different to what is done in BLEU, by employing the number of matched chunks.

\begin{comment}
\begin{align} 
\text{Penalty}&=0.5 \times ( \frac{\#\text{chunks}}{\#\text{matched unigrams}})^3,\\
\text{MEREOR}&=\frac{10PR}{R+9P}\times (1-\text{Penalty}).
\end{align}

\end{comment}

\subsubsection{\emph{Revisiting Word Order}}

%GJ: again explain why you are referring to Levenshtein distance, 
The right word order plays an important role to ensure a high quality translation output. However,  language diversity also allows different appearances or structures of a sentence. How to successfully achieve a penalty on really wrong word order, i.e. wrongly structured sentences, instead of on ``correct'' different order, i.e. the candidate sentence that has different word order to the reference, but is well structured,  has attracted a lot of interest from researchers. In fact, the Levenshtein distance (\textit{Section 3.1.1}) and n-gram based measures also contain word order information. 

Featuring the explicit assessment of word order and word choice, \newcite{WongKit2009} developed the evaluation metric ATEC (assessment of text essential characteristics). This is also based on precision and recall criteria, but with a position difference penalty coefficient attached. The word choice is assessed by matching word forms at various linguistic levels, including surface form, stem, sound and sense, and further by weighing the informativeness of each word. 

\begin{comment}
to add back: Combining the precision, order, and recall information together, \newcite{ChenKuhnLarkin2012} develop an automatic evaluation metric PORT that is initially for the tuning of the MT systems to output higher quality translation. 
Meanwhile, LEPOR...
\end{comment}

Partially inspired by this, our  work LEPOR \cite{HanWongChao2012} %to-add-back:HanWongChaoHeLu2014
is designed as a combination of augmented evaluation factors including $n$-gram based \textit{word order penalty} in addition to \textit{precision}, \textit{recall}, and \textit{enhanced sentence-length penalty}. The LEPOR metric (including \textit{h}LEPOR) was reported with top performance on the English-to-other (Spanish, German, French, Czech and Russian) language pairs in ACL-WMT13 metrics shared tasks for \textit{system level} evaluation \cite{HanWongChaoLuHeWangZhou2013}. The n-gram based variant \textit{n}LEPOR \cite{HanWongChaoHeLu2014} was also analysed by MT researchers as one of the three best performing \textit{segment level} automated metrics (together with METEOR and sentBLEU-MOSES) that correlated with human judgement at a level that was not significantly outperformed by any other metrics, on Spanish-to-English, in addition to an aggregated set of overall tested language pairs \cite{DBLP:conf/naacl/GrahamBM15}.

\subsection{Deeper Linguistic Features}

Although some of the previously outlined metrics incorporate linguistic information, e.g. synonyms and stemming in METEOR and part of speech (POS) in LEPOR, the simple n-gram word surface matching methods mainly focus on the exact matches of the surface words in the output translation. 
%GJ: what is meant by "first category here", what does "first category" refer to<<Lifeng: addressed
The advantages of the metrics based on the first category (simple n-gram word matching) are that they perform well in capturing  translation fluency \cite{LoTurmuluruWu2012}, are very fast to compute and have  low cost. On the other hand, there are also some weaknesses, for instance, syntactic information is rarely considered and the underlying assumption that a good translation is one that shares the same word surface lexical choices as the reference translations is not justified semantically. Word surface lexical similarity does not adequately reflect similarity in meaning. Translation evaluation metrics that reflect meaning similarly need to be based on similarity of semantic structure and not merely flat lexical similarity.

\subsubsection{\emph{Syntactic Similarity}}

Syntactic similarity methods usually employ the features of morphological POS information, phrase categories, phrase decompositionality or sentence structure generated by  linguistic tools such as a language parser or chunker.
% \textbf{}

In grammar, a \textbf{POS} is a linguistic category of words or lexical items, which is generally defined by the syntactic or morphological behaviour of the lexical item. Common linguistic categories of lexical items include noun, verb, adjective, adverb, and preposition. To reflect the syntactic quality of automatically translated sentences,  researchers employ POS information into their evaluations. Using the IBM model one, \newcite{PopovicVilarAvramidisBurchardt2011} evaluate translation quality by calculating the similarity scores of source and target (translated) sentences without using a reference translation, based on the morphemes, 4-gram POS and lexicon probabilities. \newcite{DahlmeierLiuNg2011} developed the TESLA evaluation metrics, combining the synonyms of bilingual phrase tables and POS information in the matching task. Other similar work using POS information include \cite{GimenezMarquez2007,PopovicNey2007,HanWongChaoHeLu2014}.%to-add-back:,HanWongChaoHeLu2014

In linguistics, a \textbf{phrase} may refer to any group of words that form a constituent, and so functions as a single unit in the syntax of a sentence. To measure an MT system's performance in translating new text types, such as in what ways the system itself could be extended to deal with new text types, \newcite{PovlsenUnderwoodMusicNeville1998} carried out  work focusing on the study of an English-to-Danish MT system. The syntactic constructions are explored with more complex linguistic knowledge, such as the identification of fronted adverbial subordinate clauses and prepositional phrases. Assuming that similar grammatical structures should occur in both source and translations, \newcite{AvramidisPopovicVilarBurchardt2011} perform evaluation on source (German) and target (English) sentences employing the features of sentence length ratio, unknown words, phrase numbers including noun phrase, verb phrase and prepositional phrase. Other similar work using phrase similarity includes \cite{LiGongZhou2012} that uses noun phrases and verb phrases from chunking, \cite{EchizenArki2010} that only uses the noun phrase chunking in automatic evaluation, and \cite{HanWongChaoHeLiZhu2013} that designs a universal phrase tagset for French to English MT evaluation.

\textbf{Syntax} is the study of the principles and processes by which sentences are constructed in particular languages. To address the overall goodness of a translated \textbf{sentence's structure}, \newcite{LiuGildea2005} employ constituent labels and head-modifier dependencies from a language parser as syntactic features for MT evaluation. They compute the similarity of dependency trees. Their experiments show that adding syntactic information can improve evaluation performance, especially for predicting the fluency of translation hypotheses. Other works that use syntactic information in evaluation include \cite{LoWu2011a} and \cite{LoTurmuluruWu2012} that use an automatic shallow parser and the RED metric \cite{DBLP:conf/coling/YuWXJLL14} that applies dependency trees.

%\textbf{There are also Dependency-based MTE, e.g. RED \cite{DBLP:conf/coling/YuWXJLL14} .. introduction of RED comes here ...}

\subsubsection{\emph{Semantic Similarity}}

As a contrast to  syntactic information, which captures  overall grammaticality or sentence structure similarity, the semantic similarity of  automatic translations and the source sentences (or references) can be measured by  employing semantic features.

To capture the semantic equivalence of sentences or text fragments, \textbf{named entity} knowledge is taken from the literature on named-entity recognition, which aims to identify and classify atomic elements in a text into different entity categories \cite{MarshPerzanowski1998,GuoXuChengLi2009}. The most commonly used entity categories include the names of persons, locations, organizations and time \cite{han2013chinese}. In the MEDAR2011 evaluation campaign, one baseline system based on Moses \cite{KoehnHoangBirchBurchFederico2007} utilized an Open NLP toolkit to perform named entity detection, in addition to other packages. The low performances from the perspective of named entities causes a drop in fluency and adequacy. In the quality estimation of the MT task in WMT 2012, \cite{Buck2012} introduced features including named entity, in addition to discriminative word lexicon, neural networks, back off behavior  \cite{RaybaudLangloisSmaili2011} and edit distance. Experiments on individual features showed that, from the perspective of the increasing the correlation score with human judgments, the named entity feature contributed the most to the overall performance, in comparisons to the impacts of other features.
%GJ: "contributed nearly the most compared to the" above - this doesn't make sense, and I can't work out what is means <<Lifeng: revised

%GJ: MTE below doesn't appear to be defined anywhere <<Lifeng:revised
% mwe4mte2015
\textbf{Multi-word Expressions} (MWEs) set obstacles for MT models due to their complexity in presentation as well as idiomaticity \cite{Sag2002MWE,han-etal-2020-multimwe,han-etal-2020-alphamwe,HanJonesSmeatonBolzoni2021decomposition4mt_MWE}. To investigate the effect of MWEs in MT evaluation (MTE), \newcite{mwe4mte2015} focused on the \textit{compositionality} of noun compounds. They identify the \textbf{noun compounds} first from the system outputs and reference with Stanford parser. The matching scores of the system outputs and reference sentences are then recalculated, adding up to the Tesla metric, by considering the predicated compositionality of identified noun compound phrases. Our own recent work in this area \cite{han-etal-2020-alphamwe} provides an extensive investigation into various MT errors caused by MWEs.

\textbf{Synonyms} are words with the same or close meanings. One of the most widely used synonym databases in the NLP literature is  WordNet \cite{MillerBeckwithFellbaumGrossMiller1990}, which is an English lexical database grouping English words into sets of synonyms. WordNet classifies  words mainly into four kinds of POS categories; Noun, Verb, Adjective, and Adverb, without prepositions, determiners, etc. Synonymous words or phrases are organized using the unit of synsets. Each synset is a hierarchical structure with the words at different levels according to their semantic relations. 

\textbf{Textual entailment} is usually used as a directive relation between text fragments. If the truth of one text fragment TA follows another text fragment TB, then there is a directional relation between TA and TB (TB $\Rightarrow$ TA). Instead of the pure logical or mathematical entailment, textual entailment in natural language processing (NLP) is usually performed with a relaxed or loose definition \cite{DaganGlickmanMagnini2006}. For instance, according to text fragment TB, if it can be inferred that the text fragment TA is \textit{most likely} to be true then the relationship TB $\Rightarrow$ TA is also established. Since the relation is directive, it means that the inverse inference (TA $\Rightarrow$ TB) is not ensured to be true \cite{DaganGlickman2004}. \newcite{castillo-estrella-2012-semantic}  present a new approach for MT evaluation based on the task of ``Semantic Textual Similarity". This problem is addressed using a textual entailment engine based on WordNet semantic features. 

\textbf{Paraphrase} is to restate the meaning of a passage of text but utilizing other words, which can be seen as bidirectional textual entailment \cite{AndroutsopoulosMalakasiotis2010}. Instead of the literal translation, word by word and line by line used by meta-phrases, a paraphrase represents a dynamic equivalent. Further knowledge of paraphrases from the aspect of linguistics is introduced in the works by \cite{McKeown1979,MeteerShaked1988,BarzilayandLee2003}. \newcite{SnoverDorrSchwartzMicciulla2006} describe a new evaluation metric TER-Plus (TERp). Sequences of words in the reference are considered to be paraphrases of a sequence of words in the hypothesis if that phrase pair occurs in the TERp phrase table. 

\textbf{Semantic roles} are employed by researchers as linguistic features in  MT evaluation. To utilize  semantic roles,  sentences are usually first shallow parsed and entity tagged. Then the semantic roles are used to specify the arguments and adjuncts that occur in both the candidate translation and reference translation. For instance, the semantic roles introduced by \newcite{GimenezMarquez2007,GimenezMarquez2008} include causative agent, adverbial adjunct, directional adjunct, negation marker, and predication adjunct, etc. In a further development, \newcite{LoWu2011a,LoWu2011b} presented the MEANT metric designed to capture the predicate-argument relations as structural relations in semantic frames, which are not reflected in the flat semantic role label features in the work of \newcite{GimenezMarquez2007}. Furthermore, instead of using uniform weights, \newcite{LoTurmuluruWu2012} weight the different types of semantic roles as empirically determined by their relative importance to the adequate preservation of meaning. Generally, semantic roles account for the semantic structure of a segment and have proved effective in assessing adequacy of translation.

\textbf{Language models} are also utilized by MT evaluation researchers. A statistical language model usually assigns a probability to a sequence of words by means of a probability distribution. \newcite{GamonAueSmets2005} propose the LM-SVM, language model, and support vector machine methods investigating the possibility of evaluating MT quality and fluency in the absence of reference translations. They evaluate the performance of the system when used as a classifier for identifying highly dis-fluent and ill-formed sentences.

%to add back: \cite{MilosKhalil2014a} designed a novel sentence level MT evaluation metric BEER, which has the advantage of incorporate large number of features in a linear model to maximize the correlation with human judgments. To make smoother sentence level scores, they explored two kinds of less sparse features including ``character $n$-grams" (e.g. stem checking) and ``abstract ordering patterns" (permutation trees). They further investigated the model with more dense features such as adequacy features, fluency features and features based on \textbf{permutation trees} \cite{MilosKhalil2014b}. In the latest version, they extended the permutation-tree \cite{GildeaSattaZhang2006} into permutation-forests model \cite{MilosKhalil2014c}, and showed stable good performance on different language pairs in WMT sentence level evaluation task.

Generally, the linguistic features mentioned above, including both syntactic and semantic features, are combined in two ways, either by following a machine learning approach \cite{AlbrechtHwa2007,LeuschNey2009}, or trying to combine a wide variety of metrics in a more simple and straightforward way, such as \cite{GimenezMarquez2008,SpeciaGimenez2010,ComellesAtseriasArranzCastellon2012}.
  
\subsection{Neural Networks for TQA}

We briefly list some works that have applied deep learning and neural networks for TQA which are promising for further exploration. 
%\cite{DBLP:conf/nlpcc/MaMZWGJL16} Ma, Yvette and Liu et al. work comes here. Rohit work also. \cite{Gupta_reval:a2015} proposed ReVal, a MTE metric based on RNN. 
%`` state-of-the-art Machine Translation (MT) evaluation metrics are complex, in-volve extensive external resources (e.g. for paraphrasing) and require tuning to achieve best results. We present a simple alternative approach based on dense vec-tor spaces and recurrent neural networks (RNNs), in particular Long Short Term Memory (LSTM) networks. ForWMT-14, our new metric scores best for two out of five language pairs, and overall best and second best on all language pairs, using Spearman and Pearson correlation, respec-tively. We also show how training data is computed automatically from WMT ranks data" .
For instance, \newcite{guzman2015MTE_NN,Guzmn17MTE_NN}
use neural networks (NNs) for TQA for pair wise modeling to choose the best hypothetical translation by comparing candidate translations with a reference, integrating syntactic and semantic information into NNs.
\newcite{Gupta_reval:a2015} proposed LSTM networks based on dense vectors to conduct TQA, while \newcite{DBLP:conf/nlpcc/MaMZWGJL16} designed a new metric based on bi-directional LSTMs, which is similar to the work of \newcite{guzman2015MTE_NN} but with less complexity by allowing the evaluation of a single hypothesis with a reference, instead of a pairwise situation.

%% file: sec-perspective.tex
%sec-perspective.tex
\section{Discussion and Perspectives}

In this section, we examine several topics that can be considered for further development of MT evaluation fields.

The first aspect is that development should involve both n-gram word surface matching and the deeper linguistic features. Because natural languages are expressive and ambiguous at different levels \cite{GimenezMarquez2007}, simple n-gram word surface  similarity based metrics limit their scope to the lexical dimension and are not sufficient to ensure that two sentences convey the same meaning or not. For instance, \cite{BurchKoehnOsborne2006} and \cite{KoehnMonz06} report that simple n-gram matching metrics tend to favor  automatic statistical MT systems. If the evaluated systems belong to different types that include rule based, human aided, and statistical systems, then the simple n-gram matching metrics, such as BLEU, give a strong disagreement between these ranking results and those of the human assessors. So  deeper linguistic features are very important in the MT evaluation procedure. 

However, inappropriate utilization, or abundant or abused utilization, of linguistic features will result in limited popularity of measures incorporating linguistic features. In the future, how to utilize the linguistic features in a more accurate, flexible and simplified way, will be one challenge in MT evaluation. Furthermore, the MT evaluation from the aspects of semantic similarity is more reasonable and reaches closer to the human judgments, so it should receive more attention.

The second major aspect is that  MT quality estimation (QE) tasks  are different to traditional MT evaluation in several ways, such as extracting reference-independent features from input sentences and their translation, obtaining quality scores based on models produced from training data, predicting the quality of an unseen translated text at system run-time, filtering out sentences which are not good enough for post processing, and selecting the best translation among multiple systems. This means that with so many challenges, the topic   will continuously attract many researchers.

Thirdly, some advanced or challenging technologies that can be further applied to  MT evaluation include the deep learning models \cite{gupta-orasan-vangenabith:2015:WMT,zhang2015deep}, semantic logic form, and decipherment model.

%% file: Appendix.tex
\section*{Appendices}

%\begin{CJK*}{UTF8}{gbsn} simplified zh \end{CJK*}
%\begin{CJK*}{UTF8}{bsmi} traditional zh \end{CJK*}

\input{sec-Evaluate-TQA}

\subsection*{Appendix B: MT QE}

%\textbf{This section gives a brief introduction of the QE tasks.}
In past years, some MT evaluation methods that do not use manually created gold reference translations were proposed. These are referred to as ``Quality Estimation (QE)". Some of the related works have already been introduced in previous sections. 
The most recent quality estimation tasks can be found at WMT12 to WMT20  \cite{BurchKoehnMonzPostSoricut2012,BojarBuckBurchFedermann2013,bojar-EtAl-2014-W14-33,bojar-EtAl-2015-WMT,specia-etal-QE2018-findings,fonseca-etal-QE2019-findings,specia-etal-QE2020-findings-wmt}. These defined a novel evaluation metric that provides some advantages over the traditional ranking metrics. The DeltaAvg metric assumes that the reference test set has a number associated with each entry that represents its extrinsic value. Given these values, their metric does not need an explicit reference ranking, the way that Spearman ranking correlation does. The goal of the DeltaAvg metric is to measure how valuable a proposed ranking is according to the extrinsic values associated with the test entries.

\begin{equation} 
DeltaAvg_v[n]= \frac{\sum\limits_{k=1}^{n-1}V(S_{1,k})}{n-1}-V(S)
\end{equation}

For scoring, two task evaluation metrics were used that have traditionally been used for measuring performance in regression tasks: Mean Absolute Error (MAE) as a primary metric, and Root of Mean Squared Error (RMSE) as a secondary metric. For a given test set S with entries $s_{i},1\leqslant{i}\leqslant|S|$, $H(s_i)$ is the proposed score for entry $s_i$ (hypothesis), and $V(s_i)$ is the reference value for entry $s_i$ (gold-standard value).

\begin{eqnarray} 
\text{MAE}&=&\frac{\sum_{i-1}^{N}|H(s_i)-V(s_i)|}{N}\\
\text{RMSE}&=&\sqrt{\frac{\sum_{i-1}^{N}(H(s_i)-V(s_i))^2}{N}}
\end{eqnarray}
where $N=|S|$. Both these metrics are non-parametric, automatic and deterministic (and therefore consistent), and extrinsically interpretable.

Some further readings on MT QE are the comparison between MT evaluation and QE \newcite{SPECIA2010MTE_vs_QE} and the QE framework model QuEst \cite{specia-etal-2013-quest};
the weakly supervised approaches for quality estimation and the limitations analysis of QE Supervised Systems \cite{Moreau2013jmt-qe,moreau2014qe_coling}, and unsupervised QE models \cite{fomicheva-etal-2020TACL_QE_unsupervised}; the recent shared tasks on QE \cite{fonseca-etal-QE2019-findings,specia-etal-QE2020-findings-wmt}.

In very recent years, the two shared tasks, i.e. MT quality estimation and traditional MT evaluation metrics, have tried to integrate into each other and benefit from both knowledge. For instance, in WMT2019 shared task, there were 10 reference-less evaluation metrics which were used for the QE task, ``QE as a Metric", as well \cite{ma-etal-2019wmt-results}.

%to add back: \subsection*{Appendix C: Further Discussion}

%to-add-back: \subsection*{Appendix D: Perspective}

\subsection*{Appendix C: Mathematical Formulas}
% https://www.overleaf.com/learn/latex/aligning_equations_with_amsmath
Some mathematical formulas that are related to aforementioned metrics: 

Section 2.1.2 -  Fluency / Adequacy / Comprehension:

\begin{eqnarray}
\text{Comprehension}=\frac{\#\text{Cottect}}{6} \\
\text{Fluency}=\frac{\frac{\text{Judgment point}-1}{\text{S}-1}}{\#\text{Sentences in passage}} \\
\text{Adequacy}=\frac{\frac{\text{Judgment point}-1}{\text{S}-1}}{\#\text{Fragments in passage}} 
\end{eqnarray}

\noindent
Section 3.1.1 - Editing Distance:

\begin{comment}

\end{comment}
\begin{eqnarray}
\text{WER}=\frac{\text{substitution+insertion+deletion}}{\text{reference}_{\text{length}}}.
\end{eqnarray}

\begin{small}
\begin{equation}
\text{PER}
=1-\frac{\text{correct}-\max(0,\text{output}_{\text{length}}-\text{reference}_{\text{length}})}{\text{reference}_{\text{length}}}.
\end{equation}
\end{small}

\begin{equation} 
\text{TER}=\frac{\#\text{of edit}}{\#\text{of average reference words}}
\end{equation}

\noindent
Section 3.1.2 - Precision and Recall:

\begin{align}
\text{BLEU}&=\text{BP}\times\exp\sum_{n=1}^{N}\lambda_{n}\log{\text{Precision}_{\text{n}}},\\
\text{BP}&=
\begin{cases}
1 & \text{if } c>r,\\
e^{1-\frac{r}{c}} & \text{if } c<=r.
\end{cases}
\end{align}

\noindent
where $c$ is the total length of candidate translation, and $r$ refers to the sum of effective reference sentence length in the corpus. Bellow is from NIST metric, then F-measure, METEOR and LEPOR:

\begin{eqnarray} 
\text{Info}=\log_{2}\big(\frac{\#\text{occurrence of}~w_1,\cdots,w_{n-1}}{\#\text{occurrence of}~ w_1,\cdots,w_n}\big)
\end{eqnarray}

\begin{eqnarray}
F_{\beta}=(1+\beta^{2})\frac{PR}{R+\beta^{2}P}
\end{eqnarray}

\begin{align} 
\text{Penalty}&=LP0.5 \times ( \frac{\#\text{chunks}}{\#\text{matched unigrams}})^3\\
\text{MEREOR}&=\frac{10PR}{R+9P}\times (1-\text{Penalty})
\end{align}

\begin{align}
\text{LEPOR}&=LP \times NPosPenal \times Harmonic(\alpha R, \beta P)
\end{align}

\begin{gather*} 
\text{\textit{h}LEPOR} = {Harmonic(w_{LP}LP},
\\ w_{NPosPenal}NPosPenal, w_{HPR}HPR)
\end{gather*}

\begin{gather*} 
\text{\textit{n}LEPOR} = LP \times NPosPenal \\
\times exp(\sum_{n=1}^{N}w_nlogHPR) \end{gather*}

%\begin{align}\text{BLEU}&=\text{BP}\times\exp\sum_{n=1}^{N}\lambda_{n}\log{\text{Precision}_{\text{n}}},\

\noindent
where, in our own metric LEPOR and its variations, \textit{n}LEPOR (\textit{n}-gram \textit{precision} and \textit{recall} LEPOR) and \textit{h}LEPOR (\textit{harmonic} LEPOR), \textit{P} and \textit{R} are for precision and recall, \textit{LP} for length penalty, \textit{NPosPenal} for n-gram position difference penalty, and \textit{HPR} for harmonic mean of precision and recall, respectively \cite{HanWongChao2012,han2013language,han2014lepor,HanWongChaoHeLu2014}.

%% file: sec-Evaluate-TQA.tex
\subsection*{Appendix A: Evaluating TQA}

%\section{Evaluating QA methods-> appendix?}
\subsection*{A.1: Statistical Significance}
If different MT systems produce translations with different qualities on a dataset, how can we ensure that they indeed own different system quality? To explore this problem, \newcite{Koehn2004} presents an investigation statistical significance testing for MT evaluation. The bootstrap re-sampling method is used to compute the statistical significance intervals for evaluation metrics on small test sets. Statistical significance usually refers to two separate notions, one of which is the p-value, the probability that the observed data will occur by chance in a given single null hypothesis. The other one is the ``Type I" error rate of a statistical hypothesis test, which is also called ``false positive" and measured by the probability of incorrectly rejecting a given null hypothesis in favour of a second alternative hypothesis \cite{Hald1998}.

\subsection*{A.2: Evaluating Human Judgment}

Since human judgments are usually trusted as the gold standards that automatic MT evaluation metrics should try to approach, the reliability and coherence of human judgments is very important. Cohen's kappa agreement coefficient is one of the most commonly used evaluation methods \cite{Cohen1960}. For the problem of nominal scale agreement between two judges, there are two relevant quantities $p_{0}$ and $p_{c}$. $p_{0}$ is the proportion of units in which the judges agreed and $p_{c}$ is the proportion of units for which agreement is expected by chance. The coefficient $k$ is simply the proportion of chance-expected disagreements which do not occur, or alternatively, it is the proportion of agreement after chance agreement is removed from consideration:

\begin{equation} 
k = \frac{p_0-p_c}{1-p_c}
\end{equation}

\noindent where $p_0-p_c$ represents the proportion of the cases in which beyond-chance agreement occurs and is the numerator of the coefficient \cite{LandisKoch1977}.

\subsection*{A.3: Correlating Manual and Automatic Score}

In this section, we introduce three correlation coefficient algorithms that have been widely used at recent WMT workshops to measure the closeness of automatic evaluation and manual judgments. The choice of correlation algorithm depends on whether scores or ranks schemes are utilized.

\subsubsection*{\emph{Pearson Correlation}}

Pearson's correlation coefficient \cite{Pearson1900} is commonly represented by the Greek letter $\rho$. The correlation between random variables X and Y denoted as $\rho_{XY}$ is measured as follows \cite{MontgomeryRunger2003}.

\begin{equation} 
\rho_{XY} = \frac{cov(X,Y)}{\sqrt{V(X)V(Y)}}=\frac{\sigma_{XY}}{\sigma_X\sigma_Y}
\end{equation}

Because the standard deviations of variable $X$ and $Y$ are higher than 0 ($\sigma_X>0$ and $\sigma_Y>0$), if the covariance $\sigma_{XY}$ between $X$ and $Y$ is positive, negative or zero, the correlation score between X and Y will correspondingly result in positive, negative or zero, respectively. Based on a sample of paired data $(X, Y)$ as $(x_i,y_i), i=1\,to\,n$ , the Pearson correlation coefficient is calculated as:

\begin{equation} \rho_{XY} = \frac{\sum_{i=1}^{n}(x_i-\mu_x)(y_i-\mu_y)}{\sqrt{\sum_{i=1}^{n}(x_i-\mu_x)^2}\sqrt{\sum_{i=1}^{n}(y_i-\mu_y)^2}}
\end{equation}

\noindent where $\mu_x$ and $\mu_y$ specify the means of discrete random variable $X$ and $Y$ respectively.

\subsubsection*{\emph{Spearman rank Correlation}}

Spearman rank correlation coefficient, a simplified version of Pearson correlation coefficient, is another algorithm to measure the correlations of automatic evaluation and manual judges, e.g. in WMT metrics task \cite{BurchKoehnMonzSchroeder2008,BurchKoehnMonzSchroeder2009,BurchKoehnMonzPeterson2010,BurchKoehnMonzZaidan2011}. When there are no ties, Spearman rank correlation coefficient, which is sometimes specified as (rs) is calculated as:

\begin{equation} 
rs_{\varphi(XY)} = 1- \frac{6\sum_{i=1}^{n}d_{i}^2}{n(n^2-1)}
\end{equation}

\noindent where $d_i$ is the difference-value (D-value) between the two corresponding rank variables $(x_i-y_i)$  in $\vec{X}=\{x_1,x_2,...,x_n\}$  and $\vec{Y}=\{y_1,y_2,...,y_n\}$  describing the system $\varphi$.

\subsubsection*{\emph{Kendall's $\tau$}}

Kendall's $\tau$ \cite{Kendall1938} has been used in recent years for the correlation between automatic order and reference order \cite{BurchKoehnMonzPeterson2010,BurchKoehnMonzZaidan2011,BurchKoehnMonzPostSoricut2012}. It is defined as:

\begin{equation} 
\tau=\frac{\text{num concordant pairs}-\text{num discordant pairs}}{\text{total pairs}}
\end{equation}

% \begin{equation} \tau=\frac{num\, concordant\, pairs\,-\, num\, discordant \, pairs}{total\, pairs}

The latest version of Kendall's $\tau$ is introduced in \cite{KendallGibbons1990}. \newcite{LebanonLafferty2002} give an overview work for Kendall's $\tau$  showing its application in calculating how much the system orders differ from the reference order. More concretely, \newcite{Lapata2003} proposed the use of Kendall's $\tau$, a measure of rank correlation, to estimate the distance between a system-generated and a human-generated gold-standard order.

\subsection*{A.4: Metrics Comparison}

There are researchers who did some work about the comparisons of different types of metrics. For example, \newcite{callison2006re,callison2007meta,lavie2013automated} mentioned that, through some qualitative analysis on some standard data set, BLEU cannot reflect MT system performance well in many situations, i.e. higher BLEU score cannot ensure better translation outputs. There are some recently developed metrics that can perform much better than the traditional ones especially on challenging sentence-level evaluation, though they are not popular yet such as nLEPOR and SentBLEU-Moses \cite{DBLP:conf/naacl/GrahamBM15,DBLP:conf/naacl/GrahamL16}. Such comparison will help MT researchers to select th appropriate metrics to use for specialist tasks.

%METEOR, NLEPOR and SENTBLEUMOSES
%\textbf{comes here \cite{DBLP:conf/naacl/GrahamBM15}\cite{DBLP:conf/naacl/GrahamL16} }

%% file: Main_MoTra21.bbl
\begin{thebibliography}{142}
\expandafter\ifx\csname natexlab\endcsname\relax\def\natexlab#1{#1}\fi

\bibitem[{Albrecht and Hwa(2007)}]{AlbrechtHwa2007}
J.~Albrecht and R.~Hwa. 2007.
\newblock A re-examination of machine learning approaches for sentence-level mt
  evaluation.
\newblock In \emph{Proceedings of the 45th Annual Meeting of the ACL, Prague,
  Czech Republic}.

\bibitem[{Androutsopoulos and
  Malakasiotis(2010)}]{AndroutsopoulosMalakasiotis2010}
Jon Androutsopoulos and Prodromos Malakasiotis. 2010.
\newblock A survey of paraphrasing and textual entailment methods.
\newblock \emph{Journal of Artificial Intelligence Research}, 38:135--187.

\bibitem[{Arnold(2003)}]{Arnold2003}
D.~Arnold. 2003.
\newblock \emph{Computers and Translation: A translator's guide-Chap8 Why
  translation is difficult for computers}.
\newblock Benjamins Translation Library.

\bibitem[{Avramidis et~al.(2011)Avramidis, Popovic, Vilar, and
  Burchardt}]{AvramidisPopovicVilarBurchardt2011}
Eleftherios Avramidis, Maja Popovic, David Vilar, and Aljoscha Burchardt. 2011.
\newblock Evaluate with confidence estimation: Machine ranking of translation
  outputs using grammatical features.
\newblock In \emph{Proceedings of WMT 2011}.

\bibitem[{Banerjee and Lavie(2005)}]{BanerjeeLavie2005}
Satanjeev Banerjee and Alon Lavie. 2005.
\newblock Meteor: An automatic metric for mt evaluation with improved
  correlation with human judgments.
\newblock In \emph{Proceedings of the ACL 2005}.

\bibitem[{Bangalore et~al.(2000)Bangalore, Rambow, and
  Whittaker}]{BangaloreRambowWhittaker2000}
Srinivas Bangalore, Owen Rambow, and Steven Whittaker. 2000.
\newblock Evaluation metrics for generation.
\newblock In \emph{Proceedings of INLG 2000}.

\bibitem[{Barrault et~al.(2020)Barrault, Biesialska, Bojar, Costa-juss{\`a},
  Federmann, Graham, Grundkiewicz, Haddow, Huck, Joanis, Kocmi, Koehn, Lo,
  Ljube{\v{s}}i{\'c}, Monz, Morishita, Nagata, Nakazawa, Pal, Post, and
  Zampieri}]{barrault-etal-wmt2020_findings}
Lo{\"\i}c Barrault, Magdalena Biesialska, Ond{\v{r}}ej Bojar, Marta~R.
  Costa-juss{\`a}, Christian Federmann, Yvette Graham, Roman Grundkiewicz,
  Barry Haddow, Matthias Huck, Eric Joanis, Tom Kocmi, Philipp Koehn, Chi-kiu
  Lo, Nikola Ljube{\v{s}}i{\'c}, Christof Monz, Makoto Morishita, Masaaki
  Nagata, Toshiaki Nakazawa, Santanu Pal, Matt Post, and Marcos Zampieri. 2020.
\newblock \href {https://www.aclweb.org/anthology/2020.wmt-1.1} {Findings of
  the 2020 conference on machine translation ({WMT}20)}.
\newblock In \emph{Proceedings of the Fifth Conference on Machine Translation},
  pages 1--55, Online. Association for Computational Linguistics.

\bibitem[{Barrault et~al.(2019)Barrault, Bojar, Costa-juss{\`a}, Federmann,
  Fishel, Graham, Haddow, Huck, Koehn, Malmasi, Monz, M{\"u}ller, Pal, Post,
  and Zampieri}]{barrault-etal-wmt2019-findings}
Lo{\"\i}c Barrault, Ond{\v{r}}ej Bojar, Marta~R. Costa-juss{\`a}, Christian
  Federmann, Mark Fishel, Yvette Graham, Barry Haddow, Matthias Huck, Philipp
  Koehn, Shervin Malmasi, Christof Monz, Mathias M{\"u}ller, Santanu Pal, Matt
  Post, and Marcos Zampieri. 2019.
\newblock \href {https://doi.org/10.18653/v1/W19-5301} {Findings of the 2019
  conference on machine translation ({WMT}19)}.
\newblock In \emph{Proceedings of the Fourth Conference on Machine Translation
  (Volume 2: Shared Task Papers, Day 1)}, pages 1--61, Florence, Italy.
  Association for Computational Linguistics.

\bibitem[{Barzilay and Lee(2003)}]{BarzilayandLee2003}
Regina Barzilay and Lillian Lee. 2003.
\newblock Learning to paraphrase: an unsupervised approach using
  multiple-sequence alignment.
\newblock In \emph{Proceedings of NAACL 2003}.

\bibitem[{Bhandari et~al.(2020)Bhandari, Gour, Ashfaq, Liu, and
  Neubig}]{bhandari-etal-2020-evaluating_summarization}
Manik Bhandari, Pranav~Narayan Gour, Atabak Ashfaq, Pengfei Liu, and Graham
  Neubig. 2020.
\newblock \href {https://doi.org/10.18653/v1/2020.emnlp-main.751}
  {Re-evaluating evaluation in text summarization}.
\newblock In \emph{Proceedings of the 2020 Conference on Empirical Methods in
  Natural Language Processing (EMNLP)}, pages 9347--9359, Online. Association
  for Computational Linguistics.

\bibitem[{Bojar et~al.(2013)Bojar, Buck, Callison-Burch, Federmann, Haddow,
  Koehn, Monz, Post, Soricut, and Specia}]{BojarBuckBurchFedermann2013}
Ond$\check{r}$ej Bojar, Christian Buck, Chris Callison-Burch, Christian
  Federmann, Barry Haddow, Philipp Koehn, Christof Monz, Matt Post, Radu
  Soricut, and Lucia Specia. 2013.
\newblock Findings of the 2013 workshop on statistical machine translation.
\newblock In \emph{Proceedings of WMT 2013}.

\bibitem[{Bojar et~al.(2014)Bojar, Buck, Federmann, Haddow, Koehn, Leveling,
  Monz, Pecina, Post, Saint-Amand, Soricut, Specia, and
  Tamchyna}]{bojar-EtAl-2014-W14-33}
Ondrej Bojar, Christian Buck, Christian Federmann, Barry Haddow, Philipp Koehn,
  Johannes Leveling, Christof Monz, Pavel Pecina, Matt Post, Herve Saint-Amand,
  Radu Soricut, Lucia Specia, and Ale\v{s} Tamchyna. 2014.
\newblock \href {http://www.aclweb.org/anthology/W/W14/W14-3302} {Findings of
  the 2014 workshop on statistical machine translation}.
\newblock In \emph{Proceedings of the {N}inth {W}orkshop on {S}tatistical
  {M}achine {T}ranslation}, pages 12--58, Baltimore, Maryland, USA. Association
  for Computational Linguistics.

\bibitem[{Bojar et~al.(2017)Bojar, Chatterjee, Federmann, Graham, Haddow,
  Huang, Huck, Koehn, Liu, Logacheva, Monz, Negri, Post, Rubino, Specia, and
  Turchi}]{wmt2017findings}
Ond{\v{r}}ej Bojar, Rajen Chatterjee, Christian Federmann, Yvette Graham, Barry
  Haddow, Shujian Huang, Matthias Huck, Philipp Koehn, Qun Liu, Varvara
  Logacheva, Christof Monz, Matteo Negri, Matt Post, Raphael Rubino, Lucia
  Specia, and Marco Turchi. 2017.
\newblock \href {https://doi.org/10.18653/v1/W17-4717} {Findings of the 2017
  conference on machine translation ({WMT}17)}.
\newblock In \emph{Proceedings of the Second Conference on Machine
  Translation}, pages 169--214, Copenhagen, Denmark. Association for
  Computational Linguistics.

\bibitem[{Bojar et~al.(2018)Bojar, Federmann, Fishel, Graham, Haddow, Huck,
  Koehn, and Monz}]{wmt2018findings}
Ond{\v{r}}ej Bojar, Christian Federmann, Mark Fishel, Yvette Graham, Barry
  Haddow, Matthias Huck, Philipp Koehn, and Christof Monz. 2018.
\newblock \href {http://www.aclweb.org/anthology/W18-6401} {Findings of the
  2018 conference on machine translation (wmt18)}.
\newblock In \emph{Proceedings of the Third Conference on Machine Translation,
  Volume 2: Shared Task Papers}, pages 272--307, Belgium, Brussels. Association
  for Computational Linguistics.

\bibitem[{Bojar et~al.(2016)Bojar, Graham, Kamran, and
  Stanojevi{\'c}}]{bojar-etal-wmt_metrics2016-results}
Ond{\v{r}}ej Bojar, Yvette Graham, Amir Kamran, and Milo{\v{s}} Stanojevi{\'c}.
  2016.
\newblock \href {https://doi.org/10.18653/v1/W16-2302} {Results of the {WMT}16
  metrics shared task}.
\newblock In \emph{Proceedings of the First Conference on Machine Translation:
  Volume 2, Shared Task Papers}, pages 199--231, Berlin, Germany. Association
  for Computational Linguistics.

\bibitem[{Bojar et~al.(2015)Bojar, Chatterjee, Federmann, Haddow, Huck, Hokamp,
  Koehn, Logacheva, Monz, Negri, Post, Scarton, Specia, and
  Turchi}]{bojar-EtAl-2015-WMT}
Ond\v{r}ej Bojar, Rajen Chatterjee, Christian Federmann, Barry Haddow, Matthias
  Huck, Chris Hokamp, Philipp Koehn, Varvara Logacheva, Christof Monz, Matteo
  Negri, Matt Post, Carolina Scarton, Lucia Specia, and Marco Turchi. 2015.
\newblock \href {http://aclweb.org/anthology/W15-3001} {Findings of the 2015
  workshop on statistical machine translation}.
\newblock In \emph{Proceedings of the {T}enth {W}orkshop on {S}tatistical
  {M}achine {T}ranslation}, pages 1--46, Lisbon, Portugal. Association for
  Computational Linguistics.

\bibitem[{Brown et~al.(1993)Brown, Della~Pietra, Della~Pietra, and
  Mercer}]{brown-etal-1993-mathematics}
Peter~F. Brown, Stephen~A. Della~Pietra, Vincent~J. Della~Pietra, and Robert~L.
  Mercer. 1993.
\newblock \href {https://www.aclweb.org/anthology/J93-2003} {The mathematics of
  statistical machine translation: Parameter estimation}.
\newblock \emph{Computational Linguistics}, 19(2):263--311.

\bibitem[{Buck(2012)}]{Buck2012}
Christian Buck. 2012.
\newblock Black box features for the wmt 2012 quality estimation shared task.
\newblock In \emph{Proceedings of WMT 2012}.

\bibitem[{Callison-Burch et~al.(2007{\natexlab{a}})Callison-Burch, Fordyce,
  Koehn, Monz, and Schroeder}]{BurchFordyceKoehnMonz2007}
Chris Callison-Burch, Cameron Fordyce, Philipp Koehn, Christof Monz, and Josh
  Schroeder. 2007{\natexlab{a}}.
\newblock (meta-) evaluation of machine translation.
\newblock In \emph{Proceedings of WMT 2007}.

\bibitem[{Callison-Burch et~al.(2007{\natexlab{b}})Callison-Burch, Fordyce,
  Koehn, Monz, and Schroeder}]{callison2007meta}
Chris Callison-Burch, Cameron Fordyce, Philipp Koehn, Christof Monz, and Josh
  Schroeder. 2007{\natexlab{b}}.
\newblock (meta-) evaluation of machine translation.
\newblock In \emph{Proceedings of the Second {W}orkshop on {S}tatistical
  {M}achine {T}ranslation}, pages 64--71. Association for Computational
  Linguistics.

\bibitem[{Callison-Burch et~al.(2008)Callison-Burch, Fordyce, Koehn, Monz, and
  Schroeder}]{BurchKoehnMonzSchroeder2008}
Chris Callison-Burch, Cameron Fordyce, Philipp Koehn, Christof Monz, and Josh
  Schroeder. 2008.
\newblock Further meta-evaluation of machine translation.
\newblock In \emph{Proceedings of WMT 2008}.

\bibitem[{Callison-Burch et~al.(2010)Callison-Burch, Koehn, Monz, Peterson,
  Przybocki, and Zaridan}]{BurchKoehnMonzPeterson2010}
Chris Callison-Burch, Philipp Koehn, Christof Monz, Kay Peterson, Mark
  Przybocki, and Omar~F. Zaridan. 2010.
\newblock Findings of the 2010 joint workshop on statistical machine
  translation and metrics for machine translation.
\newblock In \emph{Proceedings of the WMT 2010}.

\bibitem[{Callison-Burch et~al.(2012)Callison-Burch, Koehn, Monz, Post,
  Soricut, and Specia}]{BurchKoehnMonzPostSoricut2012}
Chris Callison-Burch, Philipp Koehn, Christof Monz, Matt Post, Radu Soricut,
  and Lucia Specia. 2012.
\newblock Findings of the 2012 workshop on statistical machine translation.
\newblock In \emph{Proceedings of WMT 2012}.

\bibitem[{Callison-Burch et~al.(2009)Callison-Burch, Koehn, Monz, and
  Schroeder}]{BurchKoehnMonzSchroeder2009}
Chris Callison-Burch, Philipp Koehn, Christof Monz, and Josh Schroeder. 2009.
\newblock Findings of the 2009 workshop on statistical machine translation.
\newblock In \emph{Proceedings of the 4th WMT 2009}.

\bibitem[{Callison-Burch et~al.(2011)Callison-Burch, Koehn, Monz, and
  Zaridan}]{BurchKoehnMonzZaidan2011}
Chris Callison-Burch, Philipp Koehn, Christof Monz, and Omar~F. Zaridan. 2011.
\newblock Findings of the 2011 workshop on statistical machine translation.
\newblock In \emph{Proceedings of WMT 2011}.

\bibitem[{Callison-Burch et~al.(2006{\natexlab{a}})Callison-Burch, Koehn, and
  Osborne}]{BurchKoehnOsborne2006}
Chris Callison-Burch, Philipp Koehn, and Miles Osborne. 2006{\natexlab{a}}.
\newblock Improved statistical machine translation using paraphrases.
\newblock In \emph{Proceedings of HLT-NAACL 2006}.

\bibitem[{Callison-Burch et~al.(2006{\natexlab{b}})Callison-Burch, Osborne, and
  Koehn}]{callison2006re}
Chris Callison-Burch, Miles Osborne, and Philipp Koehn. 2006{\natexlab{b}}.
\newblock Re-evaluating the role of bleu in machine translation research.
\newblock In \emph{Proceedings of EACL 2006}, volume 2006, pages 249--256.

\bibitem[{Carroll(1966)}]{Carroll1966}
John~B. Carroll. 1966.
\newblock An experiment in evaluating the quality of translation.
\newblock \emph{Mechanical Translation and Computational Linguistics},
  9(3-4):67--75.

\bibitem[{Castillo and Estrella(2012)}]{castillo-estrella-2012-semantic}
Julio Castillo and Paula Estrella. 2012.
\newblock \href {https://www.aclweb.org/anthology/W12-3103} {Semantic textual
  similarity for {MT} evaluation}.
\newblock In \emph{Proceedings of the {S}eventh {W}orkshop on {S}tatistical
  {M}achine {T}ranslation}, pages 52--58, Montr{\'e}al, Canada. Association for
  Computational Linguistics.

\bibitem[{Chiang(2005)}]{chiang-2005-hierarchical}
David Chiang. 2005.
\newblock \href {https://doi.org/10.3115/1219840.1219873} {A hierarchical
  phrase-based model for statistical machine translation}.
\newblock In \emph{Proceedings of the 43rd Annual Meeting of the Association
  for Computational Linguistics ({ACL}{'}05)}, pages 263--270, Ann Arbor,
  Michigan. Association for Computational Linguistics.

\bibitem[{Cho et~al.(2014)Cho, van Merrienboer, Bahdanau, and
  Bengio}]{cho2014encoder-decoder}
KyungHyun Cho, Bart van Merrienboer, Dzmitry Bahdanau, and Yoshua Bengio. 2014.
\newblock \href {http://arxiv.org/abs/1409.1259} {On the properties of neural
  machine translation: Encoder-decoder approaches}.
\newblock \emph{CoRR}, abs/1409.1259.

\bibitem[{Church and Hovy(1991)}]{ChurchHovy1991}
Kenneth Church and Eduard Hovy. 1991.
\newblock Good applications for crummy machine translation.
\newblock In \emph{Proceedings of the Natural Language Processing Systems
  Evaluation Workshop}.

\bibitem[{Cohen(1960)}]{Cohen1960}
Jasob Cohen. 1960.
\newblock A coefficient of agreement for nominal scales.
\newblock \emph{Educational and Psychological Measurement}, 20(1):37--46.

\bibitem[{Comelles et~al.(2012)Comelles, Atserias, Arranz, and
  Castell$\acute{o}$n}]{ComellesAtseriasArranzCastellon2012}
Elisabet Comelles, Jordi Atserias, Victoria Arranz, and Irene
  Castell$\acute{o}$n. 2012.
\newblock Verta: Linguistic features in mt evaluation.
\newblock In \emph{LREC}, pages 3944--3950.

\bibitem[{Dagan and Glickman(2004)}]{DaganGlickman2004}
Ido Dagan and Oren Glickman. 2004.
\newblock Probabilistic textual entailment: Generic applied modeling of
  language variability.
\newblock In \emph{Learning Methods for Text Understanding and Mining
  workshop}.

\bibitem[{Dagan et~al.(2006)Dagan, Glickman, and
  Magnini}]{DaganGlickmanMagnini2006}
Ido Dagan, Oren Glickman, and Bernardo Magnini. 2006.
\newblock The pascal recognising textual entailment challenge.
\newblock \emph{Machine Learning Challenges:LNCS}, 3944:177--190.

\bibitem[{Dahlmeier et~al.(2011)Dahlmeier, Liu, and Ng}]{DahlmeierLiuNg2011}
Daniel Dahlmeier, Chang Liu, and Hwee~Tou Ng. 2011.
\newblock Tesla at wmt2011: Translation evaluation and tunable metric.
\newblock In \emph{Proceedings of WMT 2011}.

\bibitem[{Doddington(2002)}]{Doddington2002}
George Doddington. 2002.
\newblock Automatic evaluation of machine translation quality using n-gram
  co-occurrence statistics.
\newblock In \emph{HLT Proceedings}.

\bibitem[{Dorr et~al.(2009)Dorr, Snover, and Nitin~Madnani}]{GALEprogram2009}
Bonnie Dorr, Matt Snover, and etc. Nitin~Madnani. 2009.
\newblock Part 5: Machine translation evaluation.
\newblock In \emph{Bonnie Dorr edited DARPA GALE program report}.

\bibitem[{Doyon et~al.(1999)Doyon, White, and
  Taylor}]{Doyon99task-basedevaluation}
Jennifer~B. Doyon, John~S. White, and Kathryn~B. Taylor. 1999.
\newblock Task-based evaluation for machine translation.
\newblock In \emph{Proceedings of MT Summit 7}.

\bibitem[{Echizen-ya and Araki(2010)}]{EchizenArki2010}
H.~Echizen-ya and K.~Araki. 2010.
\newblock Automatic evaluation method for machine translation using noun-phrase
  chunking.
\newblock In \emph{Proceedings of the ACL 2010}.

\bibitem[{Eck and Hori(2005)}]{EckHori2005}
Matthias Eck and Chiori Hori. 2005.
\newblock Overview of the iwslt 2005 evaluation campaign.
\newblock In \emph{In proceeding of International Workshop on Spoken Language
  Translation (IWSLT)}.

\bibitem[{EuroMatrix(2007)}]{EuroMatrixProject2007}
Project EuroMatrix. 2007.
\newblock 1.3: Survey of machine translation evaluation.
\newblock In \emph{EuroMatrix Project Report, Statistical and Hybrid MT between
  All European Languages, co-ordinator: Prof. Hans Uszkoreit}.

\bibitem[{Federico et~al.(2011)Federico, Bentivogli, Paul, and
  St$\ddot{u}$ker}]{FedericoBentivogliPaulStuker2011}
Marcello Federico, Luisa Bentivogli, Michael Paul, and Sebastian
  St$\ddot{u}$ker. 2011.
\newblock Overview of the iwslt 2011 evaluation campaign.
\newblock In \emph{In proceeding of International Workshop on Spoken Language
  Translation (IWSLT)}.

\bibitem[{Fomicheva et~al.(2020)Fomicheva, Sun, Yankovskaya, Blain, Guzm{\'a}n,
  Fishel, Aletras, Chaudhary, and
  Specia}]{fomicheva-etal-2020TACL_QE_unsupervised}
Marina Fomicheva, Shuo Sun, Lisa Yankovskaya, Fr{\'e}d{\'e}ric Blain, Francisco
  Guzm{\'a}n, Mark Fishel, Nikolaos Aletras, Vishrav Chaudhary, and Lucia
  Specia. 2020.
\newblock \href {https://doi.org/10.1162/tacl_a_00330} {Unsupervised quality
  estimation for neural machine translation}.
\newblock \emph{Transactions of the Association for Computational Linguistics},
  8:539--555.

\bibitem[{Fonseca et~al.(2019)Fonseca, Yankovskaya, Martins, Fishel, and
  Federmann}]{fonseca-etal-QE2019-findings}
Erick Fonseca, Lisa Yankovskaya, Andr{\'e} F.~T. Martins, Mark Fishel, and
  Christian Federmann. 2019.
\newblock \href {https://doi.org/10.18653/v1/W19-5401} {Findings of the {WMT}
  2019 shared tasks on quality estimation}.
\newblock In \emph{Proceedings of the Fourth Conference on Machine Translation
  (Volume 3: Shared Task Papers, Day 2)}, pages 1--10, Florence, Italy.
  Association for Computational Linguistics.

\bibitem[{Gamon et~al.(2005)Gamon, Aue, and Smets}]{GamonAueSmets2005}
Michael Gamon, Anthony Aue, and Martine Smets. 2005.
\newblock Sentence-level mt evaluation without reference translations beyond
  language modelling.
\newblock In \emph{Proceedings of EAMT}, pages 103--112.

\bibitem[{{Gehrmann} et~al.(2021){Gehrmann}, {Adewumi}, {Aggarwal}, {Sasanka
  Ammanamanchi}, {Anuoluwapo}, {Bosselut}, {Raghavi Chandu}, {Clinciu}, {Das},
  {Dhole}, {Du}, {Durmus}, {Du{\v{s}}ek}, {Emezue}, {Gangal}, {Garbacea},
  {Hashimoto}, {Hou}, {Jernite}, {Jhamtani}, {Ji}, {Jolly}, {Kale}, {Kumar},
  {Ladhak}, {Madaan}, {Maddela}, {Mahajan}, {Mahamood}, {Prasad Majumder},
  {Martins}, {McMillan-Major}, {Mille}, {van Miltenburg}, {Nadeem}, {Narayan},
  {Nikolaev}, {Niyongabo}, {Osei}, {Parikh}, {Perez-Beltrachini}, {Rao},
  {Raunak}, {Rodriguez}, {Santhanam}, {Sedoc}, {Sellam}, {Shaikh}, {Shimorina},
  {Sobrevilla Cabezudo}, {Strobelt}, {Subramani}, {Xu}, {Yang}, {Yerukola}, and
  {Zhou}}]{arxiv2021NLG}
Sebastian {Gehrmann}, Tosin {Adewumi}, Karmanya {Aggarwal}, Pawan {Sasanka
  Ammanamanchi}, Aremu {Anuoluwapo}, Antoine {Bosselut}, Khyathi {Raghavi
  Chandu}, Miruna {Clinciu}, Dipanjan {Das}, Kaustubh~D. {Dhole}, Wanyu {Du},
  Esin {Durmus}, Ond{\v{r}}ej {Du{\v{s}}ek}, Chris {Emezue}, Varun {Gangal},
  Cristina {Garbacea}, Tatsunori {Hashimoto}, Yufang {Hou}, Yacine {Jernite},
  Harsh {Jhamtani}, Yangfeng {Ji}, Shailza {Jolly}, Mihir {Kale}, Dhruv
  {Kumar}, Faisal {Ladhak}, Aman {Madaan}, Mounica {Maddela}, Khyati {Mahajan},
  Saad {Mahamood}, Bodhisattwa {Prasad Majumder}, Pedro~Henrique {Martins},
  Angelina {McMillan-Major}, Simon {Mille}, Emiel {van Miltenburg}, Moin
  {Nadeem}, Shashi {Narayan}, Vitaly {Nikolaev}, Rubungo~Andre {Niyongabo},
  Salomey {Osei}, Ankur {Parikh}, Laura {Perez-Beltrachini}, Niranjan~Ramesh
  {Rao}, Vikas {Raunak}, Juan~Diego {Rodriguez}, Sashank {Santhanam}, Jo{\~a}o
  {Sedoc}, Thibault {Sellam}, Samira {Shaikh}, Anastasia {Shimorina},
  Marco~Antonio {Sobrevilla Cabezudo}, Hendrik {Strobelt}, Nishant {Subramani},
  Wei {Xu}, Diyi {Yang}, Akhila {Yerukola}, and Jiawei {Zhou}. 2021.
\newblock \href {http://arxiv.org/abs/2102.01672} {{The GEM Benchmark: Natural
  Language Generation, its Evaluation and Metrics}}.
\newblock \emph{arXiv e-prints}, page arXiv:2102.01672.

\bibitem[{Gim$\acute{e}$ne and M$\acute{a}$rquez(2008)}]{GimenezMarquez2008}
Jes$\acute{u}$s Gim$\acute{e}$ne and Llu$\acute{i}$s M$\acute{a}$rquez. 2008.
\newblock A smorgasbord of features for automatic mt evaluation.
\newblock In \emph{Proceedings of WMT 2008}, pages 195--198.

\bibitem[{Gim$\acute{e}$nez and M$\acute{a}$rquez(2007)}]{GimenezMarquez2007}
Jes$\acute{u}$s Gim$\acute{e}$nez and Llu$\acute{i}$s M$\acute{a}$rquez. 2007.
\newblock Linguistic features for automatic evaluation of heterogenous mt
  systems.
\newblock In \emph{Proceedings of WMT 2007}.

\bibitem[{Graham et~al.(2015)Graham, Baldwin, and
  Mathur}]{DBLP:conf/naacl/GrahamBM15}
Yvette Graham, Timothy Baldwin, and Nitika Mathur. 2015.
\newblock \href {http://aclweb.org/anthology/N/N15/N15-1124.pdf} {Accurate
  evaluation of segment-level machine translation metrics}.
\newblock In \emph{{NAACL} {HLT} 2015, The 2015 Conference of the North
  American Chapter of the Association for Computational Linguistics: Human
  Language Technologies, Denver, Colorado, USA, May 31 - June 5, 2015}, pages
  1183--1191.

\bibitem[{Graham et~al.(2013)Graham, Baldwin, Moffat, and
  Zobel}]{graham-etal-2013-continuous}
Yvette Graham, Timothy Baldwin, Alistair Moffat, and Justin Zobel. 2013.
\newblock \href {https://www.aclweb.org/anthology/W13-2305} {Continuous
  measurement scales in human evaluation of machine translation}.
\newblock In \emph{Proceedings of the 7th Linguistic Annotation Workshop and
  Interoperability with Discourse}, pages 33--41, Sofia, Bulgaria. Association
  for Computational Linguistics.

\bibitem[{Graham et~al.(2016)Graham, Baldwin, Moffat, and Zobel}]{NLEDA2016}
Yvette Graham, Timothy Baldwin, Alistair Moffat, and Justin Zobel. 2016.
\newblock \href {https://doi.org/10.1017/S1351324915000339} {Can machine
  translation systems be evaluated by the crowd alone}.
\newblock \emph{Natural Language Engineering}, FirstView:1--28.

\bibitem[{Graham et~al.(2020)Graham, Haddow, and
  Koehn}]{graham-etal-2020-statistical}
Yvette Graham, Barry Haddow, and Philipp Koehn. 2020.
\newblock \href {https://doi.org/10.18653/v1/2020.emnlp-main.6} {Statistical
  power and translationese in machine translation evaluation}.
\newblock In \emph{Proceedings of the 2020 Conference on Empirical Methods in
  Natural Language Processing (EMNLP)}, pages 72--81, Online. Association for
  Computational Linguistics.

\bibitem[{Graham and Liu(2016)}]{DBLP:conf/naacl/GrahamL16}
Yvette Graham and Qun Liu. 2016.
\newblock \href {http://aclweb.org/anthology/N/N16/N16-1001.pdf} {Achieving
  accurate conclusions in evaluation of automatic machine translation metrics}.
\newblock In \emph{{NAACL} {HLT} 2016, The 2016 Conference of the North
  American Chapter of the Association for Computational Linguistics: Human
  Language Technologies, San Diego California, USA, June 12-17, 2016}, pages
  1--10.

\bibitem[{Guo et~al.(2009)Guo, Xu, Cheng, and Li}]{GuoXuChengLi2009}
Jiafeng Guo, Gu~Xu, Xueqi Cheng, and Hang Li. 2009.
\newblock Named entity recognition in query.
\newblock In \emph{Proceeding of SIGIR}.

\bibitem[{Gupta et~al.(2015{\natexlab{a}})Gupta, Orasan, and van
  Genabith}]{gupta-orasan-vangenabith:2015:WMT}
Rohit Gupta, Constantin Orasan, and Josef van Genabith. 2015{\natexlab{a}}.
\newblock \href {http://aclweb.org/anthology/W15-3047} {Machine translation
  evaluation using recurrent neural networks}.
\newblock In \emph{Proceedings of the {T}enth {W}orkshop on {S}tatistical
  {M}achine {T}ranslation}, pages 380--384, Lisbon, Portugal. Association for
  Computational Linguistics.

\bibitem[{Gupta et~al.(2015{\natexlab{b}})Gupta, Orasan, and van
  Genabith}]{Gupta_reval:a2015}
Rohit Gupta, Constantin Orasan, and Josef van Genabith. 2015{\natexlab{b}}.
\newblock Reval: A simple and effective machine translation evaluation metric
  based on recurrent neural networks.
\newblock In \emph{Proceedings of the 2015 Conference on Emperical Methods in
  Natural Language Processing}, pages 1066--1072. Association for Computational
  Linguistics, o.A.

\bibitem[{Guzm\'{a}n et~al.(2015)Guzm\'{a}n, Joty, M\`{a}rquez, and
  Nakov}]{guzman2015MTE_NN}
Francisco Guzm\'{a}n, Shafiq Joty, Llu\'{i}s M\`{a}rquez, and Preslav Nakov.
  2015.
\newblock \href {http://www.aclweb.org/anthology/P15-1078} {Pairwise neural
  machine translation evaluation}.
\newblock In \emph{Proceedings of the 53rd Annual Meeting of the Association
  for Computational Linguistics and The 7th International Joint Conference of
  the Asian Federation of Natural Language Processing ({ACL}'15)}, pages
  805--814, Beijing, China. Association for Computational Linguistics.

\bibitem[{Guzmn et~al.(2017)Guzmn, Joty, Mrquez, and Nakov}]{Guzmn17MTE_NN}
Francisco Guzmn, Shafiq Joty, Llus Mrquez, and Preslav Nakov. 2017.
\newblock \href {https://doi.org/10.1016/j.csl.2016.12.005} {Machine
  translation evaluation with neural networks}.
\newblock \emph{Comput. Speech Lang.}, 45(C):180--200.

\bibitem[{Hald(1998)}]{Hald1998}
Anders Hald. 1998.
\newblock \emph{A History of Mathematical Statistics from 1750 to 1930}.
\newblock ISBN-10: 0471179124. Wiley-Interscience; 1 edition.

\bibitem[{Han et~al.(2013{\natexlab{a}})Han, Wong, and Chao}]{han2013chinese}
Aaron L-F Han, Derek~F Wong, and Lidia~S Chao. 2013{\natexlab{a}}.
\newblock Chinese named entity recognition with conditional random fields in
  the light of chinese characteristics.
\newblock In \emph{Language Processing and Intelligent Information Systems},
  pages 57--68. Springer.

\bibitem[{Han(2014)}]{han2014lepor}
Lifeng Han. 2014.
\newblock \emph{LEPOR: An Augmented Machine Translation Evaluation Metric}.
\newblock University of Macau, Macao.

\bibitem[{{Han}(2016)}]{han2016MTE_survey}
Lifeng {Han}. 2016.
\newblock \href {http://arxiv.org/abs/1605.04515} {{Machine Translation
  Evaluation Resources and Methods: A Survey}}.
\newblock \emph{arXiv e-prints}, page arXiv:1605.04515.

\bibitem[{Han et~al.(2020{\natexlab{a}})Han, Jones, and
  Smeaton}]{han-etal-2020-alphamwe}
Lifeng Han, Gareth Jones, and Alan Smeaton. 2020{\natexlab{a}}.
\newblock \href {https://www.aclweb.org/anthology/2020.mwe-1.6} {{A}lpha{MWE}:
  Construction of multilingual parallel corpora with {MWE} annotations}.
\newblock In \emph{Proceedings of the Joint Workshop on Multiword Expressions
  and Electronic Lexicons}, pages 44--57, online. Association for Computational
  Linguistics.

\bibitem[{Han et~al.(2020{\natexlab{b}})Han, Jones, and
  Smeaton}]{han-etal-2020-multimwe}
Lifeng Han, Gareth Jones, and Alan Smeaton. 2020{\natexlab{b}}.
\newblock \href {https://www.aclweb.org/anthology/2020.lrec-1.363}
  {{M}ulti{MWE}: Building a multi-lingual multi-word expression ({MWE})
  parallel corpora}.
\newblock In \emph{Proceedings of the 12th Language Resources and Evaluation
  Conference}, pages 2970--2979, Marseille, France. European Language Resources
  Association.

\bibitem[{{Han} et~al.(2021){Han}, {Jones}, {Smeaton}, and
  {Bolzoni}}]{HanJonesSmeatonBolzoni2021decomposition4mt_MWE}
Lifeng {Han}, Gareth J.~F. {Jones}, Alan~F. {Smeaton}, and Paolo {Bolzoni}.
  2021.
\newblock \href {http://arxiv.org/abs/2104.04497} {{Chinese Character
  Decomposition for Neural MT with Multi-Word Expressions}}.
\newblock \emph{arXiv e-prints}, page arXiv:2104.04497.

\bibitem[{Han et~al.(2013{\natexlab{b}})Han, Wong, Chao, He, Lu, Xing, and
  Zeng}]{han2013language}
Lifeng Han, Derek~F. Wong, Lidia~S. Chao, Liangye He, Yi~Lu, Junwen Xing, and
  Xiaodong Zeng. 2013{\natexlab{b}}.
\newblock Language-independent model for machine translation evaluation with
  reinforced factors.
\newblock In \emph{Machine Translation Summit XIV}, pages 215--222.
  International Association for Machine Translation.

\bibitem[{Han et~al.(2012)Han, Wong, and Chao}]{HanWongChao2012}
Lifeng Han, Derek~Fai Wong, and Lidia~Sam Chao. 2012.
\newblock A robust evaluation metric for machine translation with augmented
  factors.
\newblock In \emph{Proceedings of COLING}.

\bibitem[{Han et~al.(2013{\natexlab{c}})Han, Wong, Chao, He, Li, and
  Zhu}]{HanWongChaoHeLiZhu2013}
Lifeng Han, Derek~Fai Wong, Lidia~Sam Chao, Liangeye He, Shuo Li, and Ling Zhu.
  2013{\natexlab{c}}.
\newblock Phrase tagset mapping for french and english treebanks and its
  application in machine translation evaluation.
\newblock In \emph{International Conference of the German Society for
  Computational Linguistics and Language Technology, LNAI Vol. 8105}, pages
  119--131.

\bibitem[{Han et~al.(2014)Han, Wong, Chao, He, and Lu}]{HanWongChaoHeLu2014}
Lifeng Han, Derek~Fai Wong, Lidia~Sam Chao, Liangeye He, and Yi~Lu. 2014.
\newblock Unsupervised quality estimation model for english to german
  translation and its application in extensive supervised evaluation.
\newblock In \emph{The Scientific World Journal. Issue: Recent Advances in
  Information Technology}, pages 1--12.

\bibitem[{Han et~al.(2013{\natexlab{d}})Han, Wong, Chao, Lu, He, Wang, and
  Zhou}]{HanWongChaoLuHeWangZhou2013}
Lifeng Han, Derek~Fai Wong, Lidia~Sam Chao, Yi~Lu, Liangye He, Yiming Wang, and
  Jiaji Zhou. 2013{\natexlab{d}}.
\newblock A description of tunable machine translation evaluation systems in
  wmt13 metrics task.
\newblock In \emph{Proceedings of WMT 2013}, pages 414--421.

\bibitem[{Johnson et~al.(2016)Johnson, Schuster, Le, Krikun, Wu, Chen, Thorat,
  Vi{\'{e}}gas, Wattenberg, Corrado, Hughes, and
  Dean}]{Google2016MultilingualNMT}
Melvin Johnson, Mike Schuster, Quoc~V. Le, Maxim Krikun, Yonghui Wu, Zhifeng
  Chen, Nikhil Thorat, Fernanda~B. Vi{\'{e}}gas, Martin Wattenberg, Greg
  Corrado, Macduff Hughes, and Jeffrey Dean. 2016.
\newblock \href {http://arxiv.org/abs/1611.04558} {Google's multilingual neural
  machine translation system: Enabling zero-shot translation}.
\newblock \emph{CoRR}, abs/1611.04558.

\bibitem[{Kendall(1938)}]{Kendall1938}
Maurice~G. Kendall. 1938.
\newblock A new measure of rank correlation.
\newblock \emph{Biometrika}, 30:81--93.

\bibitem[{Kendall and Gibbons(1990)}]{KendallGibbons1990}
Maurice~G. Kendall and Jean~Dickinson Gibbons. 1990.
\newblock \emph{Rank Correlation Methods}.
\newblock Oxford University Press, New York.

\bibitem[{King et~al.(2003)King, Popescu-Belis, and Hovy}]{KingbelisHovy2003}
Marrgaret King, Andrei Popescu-Belis, and Eduard Hovy. 2003.
\newblock Femti: Creating and using a framework for mt evaluation.
\newblock In \emph{Proceedings of the Machine Translation Summit IX}.

\bibitem[{Koehn(2004)}]{Koehn2004}
Philipp Koehn. 2004.
\newblock Statistical significance tests for machine translation evaluation.
\newblock In \emph{Proceedings of EMNLP}.

\bibitem[{Koehn(2010)}]{Koehn2010}
Philipp Koehn. 2010.
\newblock \emph{Statistical Machine Translation}.
\newblock Cambridge University Press.

\bibitem[{Koehn et~al.(2007)Koehn, Hoang, Birch, Callison-Burch, Federico,
  Bertoldi, Cowan, Shen, Moran, Zens, Dyer, Bojar, Constantin, and
  Herbst}]{KoehnHoangBirchBurchFederico2007}
Philipp Koehn, Hieu Hoang, Alexandra Birch, Chris Callison-Burch, Marcello
  Federico, Nicola Bertoldi, Brooke Cowan, Wade Shen, Christine Moran, Richard
  Zens, Chris Dyer, Ondrej Bojar, Alexandra Constantin, and Evan Herbst. 2007.
\newblock Moses: Open source toolkit for statistical machine translation.
\newblock In \emph{Proceedings of Conference on Association of Computational
  Linguistics}.

\bibitem[{Koehn and Monz(2006{\natexlab{a}})}]{koehnMonz2006}
Philipp Koehn and Christof Monz. 2006{\natexlab{a}}.
\newblock Manual and automatic evaluation of machine translation between
  european languages.
\newblock In \emph{Proceedings on the {W}orkshop on {S}tatistical {M}achine
  {T}ranslation}, pages 102--121, New York City. Association for Computational
  Linguistics.

\bibitem[{Koehn and Monz(2006{\natexlab{b}})}]{KoehnMonz06}
Philipp Koehn and Christof Monz. 2006{\natexlab{b}}.
\newblock Manual and automatic evaluation of machine translation between
  european languages.
\newblock In \emph{Proceedings of WMT 2006}.

\bibitem[{Lample and Conneau(2019)}]{bert4mt2019lample}
Guillaume Lample and Alexis Conneau. 2019.
\newblock \href {http://arxiv.org/abs/1901.07291} {Cross-lingual language model
  pretraining}.
\newblock \emph{CoRR}, abs/1901.07291.

\bibitem[{Landis and Koch(1977)}]{LandisKoch1977}
J.~Richard Landis and Gary~G. Koch. 1977.
\newblock The measurement of observer agreement for categorical data.
\newblock \emph{Biometrics}, 33(1):159--174.

\bibitem[{Laoudi et~al.(2006)Laoudi, Tate, and Voss}]{Laoudi06task-basedmt}
Jamal Laoudi, Ra~R. Tate, and Clare~R. Voss. 2006.
\newblock Task-based mt evaluation: From who/when/where extraction to event
  understanding.
\newblock In \emph{in Proceedings of LREC-06}, pages 2048--2053.

\bibitem[{Lapata(2003)}]{Lapata2003}
Mirella Lapata. 2003.
\newblock Probabilistic text structuring: Experiments with sentence ordering.
\newblock In \emph{Proceedings of ACL 2003}.

\bibitem[{L{\"a}ubli et~al.(2018)L{\"a}ubli, Sennrich, and
  Volk}]{laubli2018mt_human_parity}
Samuel L{\"a}ubli, Rico Sennrich, and Martin Volk. 2018.
\newblock \href {https://doi.org/10.18653/v1/D18-1512} {Has machine translation
  achieved human parity? a case for document-level evaluation}.
\newblock In \emph{Proceedings of the 2018 Conference on Empirical Methods in
  Natural Language Processing}, pages 4791--4796, Brussels, Belgium.
  Association for Computational Linguistics.

\bibitem[{Lavie(2013)}]{lavie2013automated}
Alon Lavie. 2013.
\newblock Automated metrics for mt evaluation.
\newblock \emph{Machine Translation}, 11:731.

\bibitem[{Lebanon and Lafferty(2002)}]{LebanonLafferty2002}
Guy Lebanon and John Lafferty. 2002.
\newblock Combining rankings using conditional probability models on
  permutations.
\newblock In \emph{Proceeding of the ICML}.

\bibitem[{Leusch and Ney(2009)}]{LeuschNey2009}
Gregor Leusch and Hermann Ney. 2009.
\newblock Edit distances with block movements and error rate confidence
  estimates.
\newblock \emph{Machine Translation}, 23(2-3).

\bibitem[{LI(2005)}]{Li2005}
A.~LI. 2005.
\newblock Results of the 2005 nist machine translation evaluation.
\newblock In \emph{Proceedings of WMT 2005}.

\bibitem[{Li et~al.(2012)Li, Gong, and Zhou}]{LiGongZhou2012}
Liang~You Li, Zheng~Xian Gong, and Guo~Dong Zhou. 2012.
\newblock Phrase-based evaluation for machine translation.
\newblock In \emph{Proceedings of COLING}, pages 663--672.

\bibitem[{{Liguori} et~al.(2021){Liguori}, {Al-Hossami}, {Cotroneo}, {Natella},
  {Cukic}, and {Shaikh}}]{code_generation2021arXiv}
Pietro {Liguori}, Erfan {Al-Hossami}, Domenico {Cotroneo}, Roberto {Natella},
  Bojan {Cukic}, and Samira {Shaikh}. 2021.
\newblock \href {http://arxiv.org/abs/2104.13100} {{Shellcode\_IA32: A Dataset
  for Automatic Shellcode Generation}}.
\newblock \emph{arXiv e-prints}, page arXiv:2104.13100.

\bibitem[{Lin and Hovy(2003)}]{LinHovy2003}
Chin-Yew Lin and E.~H. Hovy. 2003.
\newblock Automatic evaluation of summaries using n-gram co-occurrence
  statistics.
\newblock In \emph{Proceedings of NAACL 2003}.

\bibitem[{Lin and Och(2004)}]{LinOch2004}
Chin-Yew Lin and Franz~Josef Och. 2004.
\newblock Automatic evaluation of machine translation quality using longest
  common subsequence and skip-bigram statistics.
\newblock In \emph{Proceedings of ACL 2004}.

\bibitem[{Liu and Gildea(2005)}]{LiuGildea2005}
Ding Liu and Daniel Gildea. 2005.
\newblock Syntactic features for evaluation of machine translation.
\newblock In \emph{Proceedingsof theACL Workshop on Intrinsic and Extrinsic
  Evaluation Measures for Machine Translation and/or Summarization}.

\bibitem[{Lo et~al.(2012)Lo, Turmuluru, and Wu}]{LoTurmuluruWu2012}
Chi~Kiu Lo, Anand~Karthik Turmuluru, and Dekai Wu. 2012.
\newblock Fully automatic semantic mt evaluation.
\newblock In \emph{Proceedings of WMT 2012}.

\bibitem[{Lo and Wu(2011{\natexlab{a}})}]{LoWu2011a}
Chi~Kiu Lo and Dekai Wu. 2011{\natexlab{a}}.
\newblock Meant: An inexpensive, high- accuracy, semi-automatic metric for
  evaluating translation utility based on semantic roles.
\newblock In \emph{Proceedings of ACL 2011}.

\bibitem[{Lo and Wu(2011{\natexlab{b}})}]{LoWu2011b}
Chi~Kiu Lo and Dekai Wu. 2011{\natexlab{b}}.
\newblock Structured vs. flat semantic role representations for machine
  translation evaluation.
\newblock In \emph{Proceedings of the 5th Workshop on Syntax and Structure in
  StatisticalTranslation (SSST-5)}.

\bibitem[{Läubli et~al.(2020)Läubli, Castilho, Neubig, Sennrich, Shen, and
  Toral}]{L_ubli_2020_human_parity}
Samuel Läubli, Sheila Castilho, Graham Neubig, Rico Sennrich, Qinlan Shen, and
  Antonio Toral. 2020.
\newblock \href {https://doi.org/10.1613/jair.1.11371} {A set of
  recommendations for assessing human–machine parity in language
  translation}.
\newblock \emph{Journal of Artificial Intelligence Research}, 67.

\bibitem[{Ma et~al.(2016)Ma, Meng, Zheng, Wang, Graham, Jiang, and
  Liu}]{DBLP:conf/nlpcc/MaMZWGJL16}
Qingsong Ma, Fandong Meng, Daqi Zheng, Mingxuan Wang, Yvette Graham, Wenbin
  Jiang, and Qun Liu. 2016.
\newblock \href {https://doi.org/10.1007/978-3-319-50496-4_13} {Maxsd: {A}
  neural machine translation evaluation metric optimized by maximizing
  similarity distance}.
\newblock In \emph{Natural Language Understanding and Intelligent Applications
  - 5th {CCF} Conference on Natural Language Processing and Chinese Computing,
  {NLPCC} 2016, and 24th International Conference on Computer Processing of
  Oriental Languages, {ICCPOL} 2016, Kunming, China, December 2-6, 2016,
  Proceedings}, pages 153--161.

\bibitem[{Ma et~al.(2019)Ma, Wei, Bojar, and Graham}]{ma-etal-2019wmt-results}
Qingsong Ma, Johnny Wei, Ond{\v{r}}ej Bojar, and Yvette Graham. 2019.
\newblock \href {https://doi.org/10.18653/v1/W19-5302} {Results of the {WMT}19
  metrics shared task: Segment-level and strong {MT} systems pose big
  challenges}.
\newblock In \emph{Proceedings of the Fourth Conference on Machine Translation
  (Volume 2: Shared Task Papers, Day 1)}, pages 62--90, Florence, Italy.
  Association for Computational Linguistics.

\bibitem[{Mani(2001)}]{Mani2001SummarizationEA}
I.~Mani. 2001.
\newblock Summarization evaluation: An overview.
\newblock In \emph{NTCIR}.

\bibitem[{Marsh and Perzanowski(1998)}]{MarshPerzanowski1998}
Elaine Marsh and Dennis Perzanowski. 1998.
\newblock Muc-7 evaluation of ie technology: Overview of results.
\newblock In \emph{Proceedingsof Message Understanding Conference (MUC-7)}.

\bibitem[{McKeown(1979)}]{McKeown1979}
Kathleen~R. McKeown. 1979.
\newblock Paraphrasing using given and new information in a question-answer
  system.
\newblock In \emph{Proceedings of ACL 1979}.

\bibitem[{Meteer and Shaked(1988)}]{MeteerShaked1988}
Marie Meteer and Varda Shaked. 1988.
\newblock Microsoft research treelet translation system: Naacl 2006 europarl
  evaluation.
\newblock In \emph{Proceedings of COLING}.

\bibitem[{Miller et~al.(1990)Miller, Beckwith, Fellbaum, Gross, and
  Miller}]{MillerBeckwithFellbaumGrossMiller1990}
G.~A. Miller, R.~Beckwith, C.~Fellbaum, D.~Gross, and K.~J. Miller. 1990.
\newblock Wordnet: an on-line lexical database.
\newblock \emph{International Journal of Lexicography}, 3(4):235--244.

\bibitem[{Montgomery and Runger(2003)}]{MontgomeryRunger2003}
Douglas~C. Montgomery and George~C. Runger. 2003.
\newblock \emph{Applied statistics and probability for engineers}, third
  edition.
\newblock John Wiley and Sons, New York.

\bibitem[{Moreau and Vogel(2013)}]{Moreau2013jmt-qe}
Erwan Moreau and Carl Vogel. 2013.
\newblock \href {https://doi.org/10.1007/s10590-013-9142-8} {Weakly supervised
  approaches for quality estimation}.
\newblock \emph{Machine Translation}, 27(3–4):257–280.

\bibitem[{Moreau and Vogel(2014)}]{moreau2014qe_coling}
Erwan Moreau and Carl Vogel. 2014.
\newblock \href {https://www.aclweb.org/anthology/C14-1208} {Limitations of
  {MT} quality estimation supervised systems: The tails prediction problem}.
\newblock In \emph{Proceedings of {COLING} 2014, the 25th International
  Conference on Computational Linguistics: Technical Papers}, pages 2205--2216,
  Dublin, Ireland. Dublin City University and Association for Computational
  Linguistics.

\bibitem[{Och and Ney(2003)}]{Och2003CL}
Franz~Josef Och and Hermann Ney. 2003.
\newblock \href {https://doi.org/10.1162/089120103321337421} {A systematic
  comparison of various statistical alignment models}.
\newblock \emph{Computational Linguistics}, 29(1):19--51.

\bibitem[{Papineni et~al.(2002)Papineni, Roukos, Ward, and
  Zhu}]{PapineniRoukosWardZhu2002}
Kishore Papineni, Salim Roukos, Todd Ward, and Wei~Jing Zhu. 2002.
\newblock Bleu: a method for automatic evaluation of machine translation.
\newblock In \emph{Proceedings of ACL 2002}.

\bibitem[{Paul(2009)}]{Paul2009}
M.~Paul. 2009.
\newblock Overview of the iwslt 2009 evaluation campaign.
\newblock In \emph{Proceeding of IWSLT}.

\bibitem[{Paul et~al.(2010)Paul, Federico, and
  St$\ddot{u}$ker}]{PaulFedericoStuker2010}
Michael Paul, Marcello Federico, and Sebastian St$\ddot{u}$ker. 2010.
\newblock Overview of the iwslt 2010 evaluation campaign.
\newblock In \emph{Proceeding of IWSLT}.

\bibitem[{Pearson(1900)}]{Pearson1900}
Karl Pearson. 1900.
\newblock On the criterion that a given system of deviations from the probable
  in the case of a correlated system of variables is such that it can be
  reasonably supposed to have arisen from random sampling.
\newblock \emph{Philosophical Magazine}, 50(5):157--175.

\bibitem[{Popovi$\acute{c}$ et~al.(2011)Popovi$\acute{c}$, Vilar, Avramidis,
  and Burchardt}]{PopovicVilarAvramidisBurchardt2011}
Maja Popovi$\acute{c}$, David Vilar, Eleftherios Avramidis, and Aljoscha
  Burchardt. 2011.
\newblock Evaluation without references: Ibm1 scores as evaluation metrics.
\newblock In \emph{Proceedings of WMT 2011}.

\bibitem[{Popovic and Ney(2007)}]{PopovicNey2007}
M.~Popovic and Hermann Ney. 2007.
\newblock Word error rates: Decomposition over pos classes and applications for
  error analysis.
\newblock In \emph{Proceedings of WMT 2007}.

\bibitem[{Popovi{\'c}(2020{\natexlab{a}})}]{popovic-2020coling-informative}
Maja Popovi{\'c}. 2020{\natexlab{a}}.
\newblock \href {https://doi.org/10.18653/v1/2020.coling-main.444} {Informative
  manual evaluation of machine translation output}.
\newblock In \emph{Proceedings of the 28th International Conference on
  Computational Linguistics}, pages 5059--5069, Barcelona, Spain (Online).
  International Committee on Computational Linguistics.

\bibitem[{Popovi{\'c}(2020{\natexlab{b}})}]{popovic-2020conll-relations}
Maja Popovi{\'c}. 2020{\natexlab{b}}.
\newblock \href {https://doi.org/10.18653/v1/2020.conll-1.19} {Relations
  between comprehensibility and adequacy errors in machine translation output}.
\newblock In \emph{Proceedings of the 24th Conference on Computational Natural
  Language Learning}, pages 256--264, Online. Association for Computational
  Linguistics.

\bibitem[{Povlsen et~al.(1998)Povlsen, Underwood, Music, and
  Neville}]{PovlsenUnderwoodMusicNeville1998}
Claus Povlsen, Nancy Underwood, Bradley Music, and Anne Neville. 1998.
\newblock Evaluating text-type suitability for machine translation a case study
  on an english-danish system.
\newblock In \emph{Proceeding LREC}.

\bibitem[{Raybaud et~al.(2011)Raybaud, Langlois, and
  Sma$\ddot{i}$li}]{RaybaudLangloisSmaili2011}
Sylvain Raybaud, David Langlois, and Kamel Sma$\ddot{i}$li. 2011.
\newblock ''this sentence is wrong.'' detecting errors in machine-translated
  sentences.
\newblock \emph{Machine Translation}, 25(1):1--34.

\bibitem[{Reeder(2004)}]{reeder2004amta}
Florence Reeder. 2004.
\newblock Investigation of intelligibility judgments.
\newblock In \emph{Machine Translation: From Real Users to Research}, pages
  227--235, Berlin, Heidelberg. Springer Berlin Heidelberg.

\bibitem[{{Ruder} et~al.(2021){Ruder}, {Constant}, {Botha}, {Siddhant},
  {Firat}, {Fu}, {Liu}, {Hu}, {Neubig}, and
  {Johnson}}]{2021NLU_ruder_multilingual_evaluation}
Sebastian {Ruder}, Noah {Constant}, Jan {Botha}, Aditya {Siddhant}, Orhan
  {Firat}, Jinlan {Fu}, Pengfei {Liu}, Junjie {Hu}, Graham {Neubig}, and Melvin
  {Johnson}. 2021.
\newblock \href {http://arxiv.org/abs/2104.07412} {{XTREME-R: Towards More
  Challenging and Nuanced Multilingual Evaluation}}.
\newblock \emph{arXiv e-prints}, page arXiv:2104.07412.

\bibitem[{Sag et~al.(2002)Sag, Baldwin, Bond, Copestake, and
  Flickinger}]{Sag2002MWE}
Ivan~A. Sag, Timothy Baldwin, Francis Bond, Ann Copestake, and Dan Flickinger.
  2002.
\newblock Multiword expressions: A pain in the neck for nlp.
\newblock In \emph{Computational Linguistics and Intelligent Text Processing},
  pages 1--15, Berlin, Heidelberg. Springer Berlin Heidelberg.

\bibitem[{Salehi et~al.(2015)Salehi, Mathur, Cook, and Baldwin}]{mwe4mte2015}
Bahar Salehi, Nitika Mathur, Paul Cook, and Timothy Baldwin. 2015.
\newblock \href {https://doi.org/10.3115/v1/W15-0909} {The impact of multiword
  expression compositionality on machine translation evaluation}.
\newblock In \emph{Proceedings of the 11th Workshop on Multiword Expressions},
  pages 54--59, Denver, Colorado. Association for Computational Linguistics.

\bibitem[{Snover et~al.(2006)Snover, Dorr, Schwartz, Micciulla, and
  Makhoul}]{SnoverDorrSchwartzMicciulla2006}
Mattthew Snover, Bonnie~J. Dorr, Richard Schwartz, Linnea Micciulla, and John
  Makhoul. 2006.
\newblock A study of translation edit rate with targeted human annotation.
\newblock In \emph{Proceeding of AMTA}.

\bibitem[{Specia and Gim$\acute{e}$nez(2010)}]{SpeciaGimenez2010}
L.~Specia and J.~Gim$\acute{e}$nez. 2010.
\newblock Combining confidence estimation and reference-based metrics for
  segment-level mt evaluation.
\newblock In \emph{The Ninth Conference of the Association for Machine
  Translation in the Americas (AMTA)}.

\bibitem[{Specia et~al.(2020)Specia, Blain, Fomicheva, Fonseca, Chaudhary,
  Guzm{\'a}n, and Martins}]{specia-etal-QE2020-findings-wmt}
Lucia Specia, Fr{\'e}d{\'e}ric Blain, Marina Fomicheva, Erick Fonseca, Vishrav
  Chaudhary, Francisco Guzm{\'a}n, and Andr{\'e} F.~T. Martins. 2020.
\newblock \href {https://www.aclweb.org/anthology/2020.wmt-1.79} {Findings of
  the {WMT} 2020 shared task on quality estimation}.
\newblock In \emph{Proceedings of the Fifth Conference on Machine Translation},
  pages 743--764, Online. Association for Computational Linguistics.

\bibitem[{Specia et~al.(2018)Specia, Blain, Logacheva, F.~Astudillo, and
  Martins}]{specia-etal-QE2018-findings}
Lucia Specia, Fr{\'e}d{\'e}ric Blain, Varvara Logacheva, Ram{\'o}n
  F.~Astudillo, and Andr{\'e} F.~T. Martins. 2018.
\newblock \href {https://doi.org/10.18653/v1/W18-6451} {Findings of the {WMT}
  2018 shared task on quality estimation}.
\newblock In \emph{Proceedings of the Third Conference on Machine Translation:
  Shared Task Papers}, pages 689--709, Belgium, Brussels. Association for
  Computational Linguistics.

\bibitem[{Specia et~al.(2011)Specia, Hajlaoui, Hallett, and
  Aziz}]{SpeciaHajlaouiHallettAziz2011}
Lucia Specia, Naheh Hajlaoui, Catalina Hallett, and Wilker Aziz. 2011.
\newblock Predicting machine translation adequacy.
\newblock In \emph{Machine Translation Summit XIII}.

\bibitem[{Specia et~al.(2010)Specia, Raj, and Turchi}]{SPECIA2010MTE_vs_QE}
Lucia Specia, Dhwaj Raj, and Marco Turchi. 2010.
\newblock Machine translation evaluation versus quality estimation.
\newblock \emph{Machine translation}.

\bibitem[{Specia et~al.(2013)Specia, Shah, de~Souza, and
  Cohn}]{specia-etal-2013-quest}
Lucia Specia, Kashif Shah, Jose~G.C. de~Souza, and Trevor Cohn. 2013.
\newblock \href {https://www.aclweb.org/anthology/P13-4014} {{Q}u{E}st - a
  translation quality estimation framework}.
\newblock In \emph{Proceedings of the 51st Annual Meeting of the Association
  for Computational Linguistics: System Demonstrations}, pages 79--84, Sofia,
  Bulgaria. Association for Computational Linguistics.

\bibitem[{Su et~al.(1992)Su, Ming-Wen, and Jing-Shin}]{SuWenShin1992}
Keh-Yih Su, Wu~Ming-Wen, and Chang Jing-Shin. 1992.
\newblock A new quantitative quality measure for machine translation systems.
\newblock In \emph{Proceeding of COLING}.

\bibitem[{Tillmann et~al.(1997)Tillmann, Vogel, Ney, Zubiaga, and
  Sawaf}]{TillmannVogelNeyZubiagaSawaf1997}
Christoph Tillmann, Stephan Vogel, Hermann Ney, Arkaitz Zubiaga, and Hassan
  Sawaf. 1997.
\newblock Accelerated dp based search for statistical translation.
\newblock In \emph{Proceeding of EUROSPEECH}.

\bibitem[{Turian et~al.(2006)Turian, Shea, and Melamed}]{turian2006evaluation}
Joseph~P Turian, Luke Shea, and I~Dan Melamed. 2006.
\newblock Evaluation of machine translation and its evaluation.
\newblock Technical report, DTIC Document.

\bibitem[{Vaswani et~al.(2017)Vaswani, Shazeer, Parmar, Uszkoreit, Jones,
  Gomez, Kaiser, and Polosukhin}]{google2017attention}
Ashish Vaswani, Noam Shazeer, Niki Parmar, Jakob Uszkoreit, Llion Jones,
  Aidan~N. Gomez, Lukasz Kaiser, and Illia Polosukhin. 2017.
\newblock Attention is all you need.
\newblock In \emph{Conference on Neural Information Processing System}, pages
  6000--6010.

\bibitem[{Voss and Tate(2006)}]{Voss06task-basedevaluation}
Clare~R. Voss and Ra~R. Tate. 2006.
\newblock Task-based evaluation of machine translation (mt) engines: Measuring
  how well people extract who, when, where-type elements in mt output.
\newblock In \emph{In Proceedings of 11th Annual Conference of the European
  Association for Machine Translation (EAMT-2006}, pages 203--212.

\bibitem[{Weaver(1955)}]{Weaver1955}
Warren Weaver. 1955.
\newblock Translation.
\newblock \emph{Machine Translation of Languages: Fourteen Essays}.

\bibitem[{White et~al.(1994)White, Connell, and Mara}]{WhiteConnellMara1994}
John~S. White, Theresa~O' Connell, and Francis~O' Mara. 1994.
\newblock The arpa mt evaluation methodologies: Evolution, lessons, and future
  approaches.
\newblock In \emph{Proceeding of AMTA}.

\bibitem[{White and Taylor(1998)}]{WhiteTaylor1998}
John~S. White and Kathryn~B. Taylor. 1998.
\newblock A task-oriented evaluation metric for machine translation.
\newblock In \emph{Proceeding LREC}.

\bibitem[{Wong and yu~Kit(2009)}]{WongKit2009}
Billy Wong and Chun yu~Kit. 2009.
\newblock Atec: automatic evaluation of machine translation via word choice and
  word order.
\newblock \emph{Machine Translation}, 23(2-3):141--155.

\bibitem[{Yu et~al.(2014)Yu, Wu, Xie, Jiang, Liu, and
  Lin}]{DBLP:conf/coling/YuWXJLL14}
Hui Yu, Xiaofeng Wu, Jun Xie, Wenbin Jiang, Qun Liu, and Shouxun Lin. 2014.
\newblock \href {http://aclweb.org/anthology/C/C14/C14-1193.pdf} {{RED:} {A}
  reference dependency based {MT} evaluation metric}.
\newblock In \emph{{COLING} 2014, 25th International Conference on
  Computational Linguistics, Proceedings of the Conference: Technical Papers,
  August 23-29, 2014, Dublin, Ireland}, pages 2042--2051.

\bibitem[{Zhang and Zong(2015)}]{zhang2015deep}
Jiajun Zhang and Chengqing Zong. 2015.
\newblock Deep neural networks in machine translation: An overview.
\newblock \emph{IEEE Intelligent Systems}, (5):16--25.

\end{thebibliography}
